\documentclass[lettersize,journal]{IEEEtran}
\usepackage{amsmath,amsfonts}
\usepackage{algorithmic}
\usepackage{algorithm}
\usepackage{array}
\usepackage[caption=false,font=normalsize,labelfont=sf,textfont=sf]{subfig}
\usepackage{textcomp}
\usepackage{stfloats}
\usepackage{url}
\usepackage{verbatim}
\usepackage{graphicx}
\usepackage{cite}
\usepackage{xcolor}
\usepackage{colortbl}
\usepackage{hhline}
\usepackage{subcaption}
\usepackage{pifont}
\usepackage{tikz}
\usepackage{hyperref}
\usepackage{multirow}
\usepackage{booktabs} 
\begin{document}

\title{Mitigating Multilingual Hallucination in Large \\ Vision-Language Models}

\author{Xiaoye Qu, Mingyang Song, Wei Wei, Jianfeng Dong, Yu Cheng

\thanks{X. Qu and W. Wei are with the School of Computer Science \& Technology, Huazhong University of Science and Technology. Email: xiaoye@hust.edu.cn, weiw@hust.edu.cn.}

\thanks{M. Song is with the School of Computing Science and Technology, Fudan University. Email: mysong23@m.fudan.edu.cn.}

\thanks{D. Liu is with Wangxuan Institute of Computer Technology, Peking
University. Email: dzliu@stu.pku.edu.cn}

\thanks{J. Dong is with the School of Computing Science and Technology, Zhejiang Gongshang University. Email: dongjf24@gmail.com.}

\thanks{Y. Cheng is with the Department of Computer Science and Engineering,
The Chinese University of Hong Kong. E-mail: chengyu@cse.cuhk.edu.hk.}

\thanks{Xiaoye Qu and Mingyang Song are co-first authors. Corresponding author: Wei Wei.}}

\markboth{Journal of \LaTeX\ Class Files,~Vol.~14, No.~8, August~2021}%
{Shell \MakeLowercase{\textit{et al.}}: A Sample Article Using IEEEtran.cls for IEEE Journals}

\IEEEpubidadjcol

\maketitle

\begin{abstract}
While Large Vision-Language Models (LVLMs) have exhibited remarkable capabilities across a wide range of tasks, they suffer from hallucination problems, where models generate plausible yet incorrect answers given the input image-query pair. 
This hallucination phenomenon is even more severe when  
querying the image in non-English languages, while existing methods for mitigating hallucinations in LVLMs only consider the English scenarios. 
In this paper, we make the first attempt to mitigate this important multilingual hallucination in LVLMs. 
{With thorough experiment analysis, we found that multilingual hallucination in LVLMs is a systemic problem that could arise from deficiencies in multilingual capabilities or inadequate multimodal abilities. 
}
To this end, we propose a two-stage Multilingual Hallucination Removal (MHR) framework for LVLMs, aiming to improve resistance to hallucination for both high-resource and low-resource languages. 
Specifically, in the first stage, considering that most non-English languages can not follow instructions well and output non-sense answers given the input image, we boost multilingual instruction following ability with {a multilingual} supervised fine-tuning. 
The second phase is aimed at enhancing the LVLM's ability to diminish multilingual hallucinations. {Instead of relying on the intricate manual annotations of multilingual resources, we fully leverage the inherent capabilities of the LVLM and propose a novel cross-lingual alignment method}, which generates multiple responses for each image-query input and then identifies the hallucination-aware pairs for each language.

These data pairs are finally used for direct preference optimization to prompt the LVLMs to favor non-hallucinating responses.   
Experimental results show that our MHR achieves a substantial reduction in hallucination generation for LVLMs. Notably, on our extended multilingual POPE benchmark, our framework delivers an average increase of 19.0\% in accuracy across 13 different languages.
Our code and model weights are available at \href{https://github.com/ssmisya/MHR}{https://github.com/ssmisya/MHR}.
\end{abstract}


\section{Introduction}

\IEEEPARstart{L}{arge} Vision-Language Models (LVLMs) \cite{liu2023llava,ye2023mplug,bai2023qwen,mckinzie2024mm1,lin2024moe,wang2024visionllm,liu2024textmonkey,xu2023pointllm} have made significant strides in bridging visual and textual content, leading to notable developments in numerous downstream tasks \cite{shah2023lm,zhu2023prompt,li2024lmeye,liu2024online,kelly2024visiongpt,zhang2024vision,liu2024survey}. However, most recently proposed LVLMs suffer from the severe hallucination issues, where the model's output contains spurious information, such as non-existent objects, or inaccurate attributes or relations, posing a considerable challenge to the practical application of LVLMs. 

Recently, numerous techniques \cite{liu2023mitigating,ben2023mocha,sun2023aligning,yu2023rlhf,yu2024hallucidoctor,wang2024mitigating} adopting supervised fine-tuning (SFT) or Reinforcement Learning from Human Feedback (RLHF) have been proposed to combat hallucinations in LVLMs. 
\begin{figure}[!t]
\centering
\includegraphics[width=0.40\textwidth]{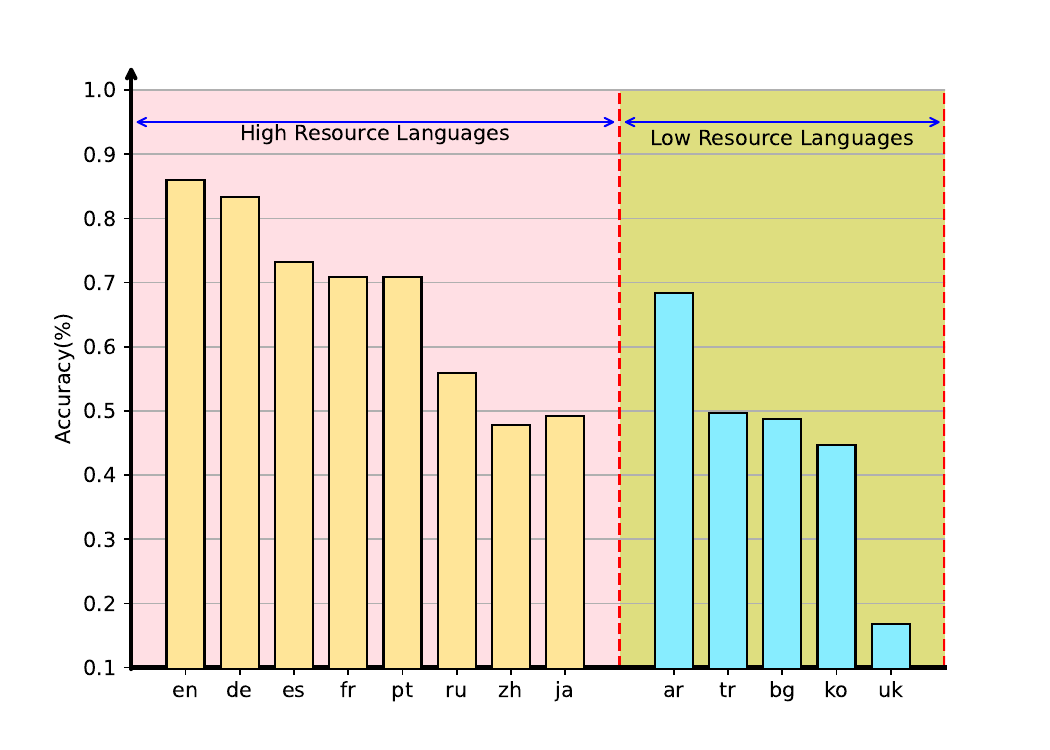}
\caption{Multilingual hallucinations in LLaVA-1.5. On the POPE MSCOCO benchmark, 
most languages have an accuracy under 70\%, but English exceeds 85\%.
} 
\label{fig:1}
\end{figure}
Given an input image and query, SFT requires an answer with no hallucination to fine-tune the model, while RLHF relies on a pair of hallucinatory and non-hallucinatory responses to learn to distinguish. Although the SFT method is beneficial, it pays more attention to the model following instructions and lacks custom hallucination handling. In contrast, RLHF performs preference optimization to promote choosing non-hallucinatory responses while rejecting answers with hallucinations. 
Considering the training difficulty of RLHF, direct optimization preference (DPO) has attracted increasing attention due to its simplicity and efficiency. HA-DPO \cite{zhao2023hadpo} builds a hallucination-aware dataset that contains both hallucination and non-hallucination responses, and effectively alleviates hallucinations in LVLMs.
However, all the above methods focus on building English datasets and have not taken into account the hallucination of multilingualism existing in LVLMs.
As shown in Figure \ref{fig:1}, on the multilingual POPE benchmark, apart from English, the accuracy of most languages is close to or less than 70\%. 
\footnote{In this paper, we consider 13 languages. If the proportion of users of a certain language exceeds 1\% of the total world population, we define it as a high-resource language, otherwise, it is a low-resource language.} 
It indicates that the hallucinations of multilingualism in LVLMs are very serious and dramatically harm the model performance for both high-resource and low-resource languages.

Therefore, in this paper, we make the first attempt to mitigate the multilingual hallucination problem in LVLMs and propose a two-stage Multilingual Hallucination Removal (MHR) framework. 
{Concretely, MHR is based on a detailed empirical analysis that unveils the underlying causalities of multilingual hallucinations in LVLMs. Our empirical findings reveal that multilingual hallucinations can be attributed to two key factors. First, except for English, most languages cannot accurately understand instructions, leading to the hallucination output of non-sense answers. 
Second, due to the lack of hallucination-aware training data for the corresponding language, these languages do not have enough ability to distinguish hallucinations, thus resulting in inaccurate answers.}

Based on the above analysis, our proposed MHR framework first improves multilingual instruction following ability with multilingual supervised fine-tuning. 
It is crucial and contributes to robust query understanding for different languages. {Bypassing this step will lead to a significant performance drop against hallucinations.}
Subsequently, we aim to improve the model's ability to resist hallucinations. {However, the hallucination-related multilingual resource is scarce and building corpus for non-English languages is time-consuming and labor-intensive. Inspired by it, we propose a novel cross-lingual alignment method and 
build multilingual hallucination-aware pairs 
by fully leveraging the inherent capabilities of the LVLM, thus avoiding manually collecting data.} 

{Specifically, to generate abundant training data, we first use the LVLM model to generate multiple multilingual responses for each query, and then identify the hallucination-aware pairs according to the cross-lingual alignment metric}. Finally, these data pairs are used for direct preference optimization to prompt
the LVLM to favor non-hallucinating responses.

To verify the effectiveness of our MHR framework, we extend the traditional benchmarks used for English hallucination evaluation, such as discriminative benchmark POPE, MME, and generative benchmark AMBER into multi-language evaluation sets POPE MUL, MME MUL, and AMBER MUL. 
The experiments on recent LVLMs LLaVA 1.5 and CogVLM demonstrate that our MHR significantly improves the model performance against hallucination for both high-resource and low-resource languages. 

To sum up, our contributions are summarized as follows:

\begin{itemize}

    \item To the best of our knowledge, this is the first work that mitigates multilingual hallucinations in LVLMs. Our work reveals the severe multilingual hallucination problem 
    when querying the existing LVLMs using non-English languages. Meanwhile, we analyze the two factors of causing multilingual hallucinations in LVLMs.

    \item To mitigate the complex multilingual hallucination in LVLMs, we propose a two-stage multilingual hallucination removal (MHR)  framework. It first improves the instruction-following ability for different languages and then enhances the ability to eliminate hallucinations with a hallucination-enhanced preference optimization.

     \item {Instead of building multilingual datasets manually, 
     we introduce a novel cross-lingual alignment method that automatically constructs multilingual datasets and 
     improves the ability of multilingual hallucination resistance. 
     During this process, the reasoning processes of other languages are aligned with those of English.}

    \item We broaden the traditional discriminative and generative hallucination benchmarks in LVLMs to encompass multilingual contexts. On these extended benchmarks, our framework delivers substantial advancements in mitigating hallucination. 
    Remarkably, our approach yields an average improvement of 19.0 points on POPE benchmark.

\end{itemize}

\section{Related Work}

\subsection{Large Visual-Language Models}

Recently, large-scale models have attracted significant attention and wide applications \cite{su2024timo,su2024living,lu2024mitigating}. 
With the aid of strong large language models such as LLaMA \cite{touvron2023llama} and Vicuna \cite{vicuna2023}, a batch of LVLMs such as LLaVA 1.5 \cite{liu2023llava,liu2023improvedllava}, InstructBLIP \cite{dai2024instructblip}, miniGPT4 \cite{zhu2023minigpt}, and CogVLM \cite{wang2023cogvlm} have emerged, which can comprehend and generate a wide array of content by utilizing information from distinct modalities like texts and images.
These models undergo two training phases: pre-trained feature alignment and instruction fine-tuning, which assist the model in comprehending instruction input formats. However, all aforementioned LVLMs still encounter significant hallucination issues. Thus, in this paper, we focus on solving hallucination problems to promote the use of LVLMs in practical scenarios.

\subsection{Hallucination in LVLMs}

In the realm of LVLMs, hallucination refers to generating output that is irrelevant or factually inaccurate given the input image-query pair. 
It is a significant issue in LVLMs that may arise due to biased training data, overfitting training data, or poor comprehension of real-world facts. 
Approaches for mitigating hallucination issues in current LVLMs mainly focus on refining the training process and building hallucination-resistant datasets. 
Yu et al. \cite{liu2023mitigating} apply robust instruct tuning to augment the model performance. Sun et al. \cite{sun2023aligning} propose aligning the model with factually augmented Reinforcement Learning with Human Feedback (RLHF), facilitating factually accurate outputs. Similar to above method, Yu et al. \cite{yu2023rlhf} also adopt RLHF but devise a fine-grained version. 
Considering the inherent complexity in model training processes when utilizing above RLHF methods to constrain model preferences, \cite{zhao2023beyond} further introduce a hallucination-aware direct preference optimization (DPO), which is both simple and efficient. 
In this paper, we adopt the direct preference optimization considering its flexibility and superior performance. 
It is worth noting that our proposed framework can be easily applied to existing LVLMs. {In the experiment, we apply our method to strong LVLM models LLaVA 1.5 and CogVLM and has verified the effectiveness}. Moreover, the above approaches proposed for mitigating hallucination only focus on their effectiveness in English. 
Although these methods might be transferable to other languages, constructing datasets for each single language is a very time-consuming and labor-intensive process. 
In this paper, we systematically analyze the causes of multilingual hallucinations and propose a novel and automatic method for multilingual dataset construction.

\begin{figure*}
\centering
\includegraphics[width=17cm]{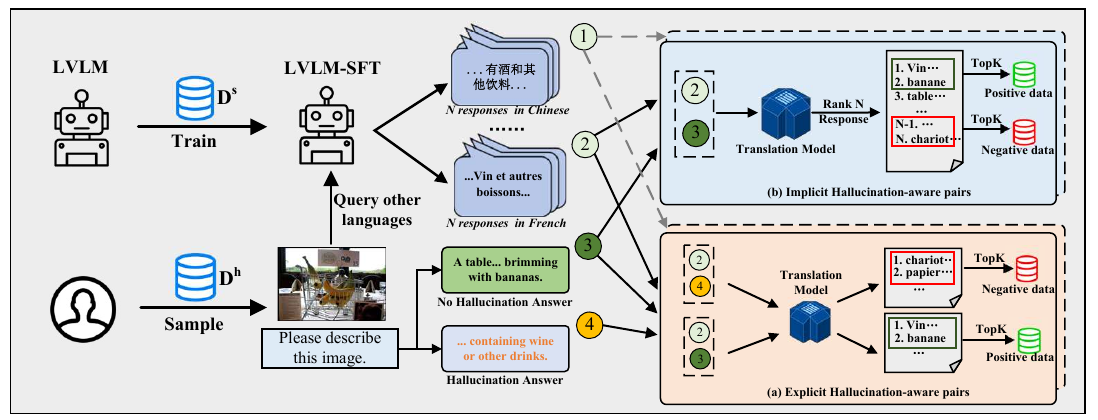}
\caption{Our Multilingual Hallucination Removal framework. Firstly, LVLM enhances the ability to follow multilingual instructions through supervised fine-tuning. Subsequently, based on an existing hallucination-aware dataset $D^h$, LVLM generates $N$ responses for each language given the corresponding language query. 
Then, the responses, English hallucination answer, and English no hallucination answers are used to generate haluciantion-aware pairs for final direct preference optimization.  
}
\label{SEN}
\end{figure*}

\subsection{Multilingual Large Vision-Language Models}

In the era of the large models, based on the stong language model LLaMA \cite{touvron2023llama} which trains on the multilingual corpus, LLaVA is an inherently multilingual vision-language model. 
Recently, PALO \cite{palo} enhances LLaVA's \cite{liu2024visual} multilingual ability and procures a large, multilingual (spanning 10 languages) instruction-tuning dataset. Unfortunately, the PALO model remains undisclosed.  
In addition, MBLIP \cite{geigle2023mblip} broadens the multilingual capability of BLIP2 \cite{li2023blip2} by recalibrating an image encoder initially aligned with an English LLM to a newly developed multilingual LLM supplemented with a diverse array of multilingual visual-language data, encompassing up to 96 languages. 
However, there are no existing works mitigating the multilingual hallucination in LVLMs. In this paper, based on the widely-used LVLMs, we make the first attempt to solve hallucinations in LVLMs for both high-resource and low-source languages, covering a total of 13 languages.

\begin{figure}
\centering
\includegraphics[width=0.46\textwidth]{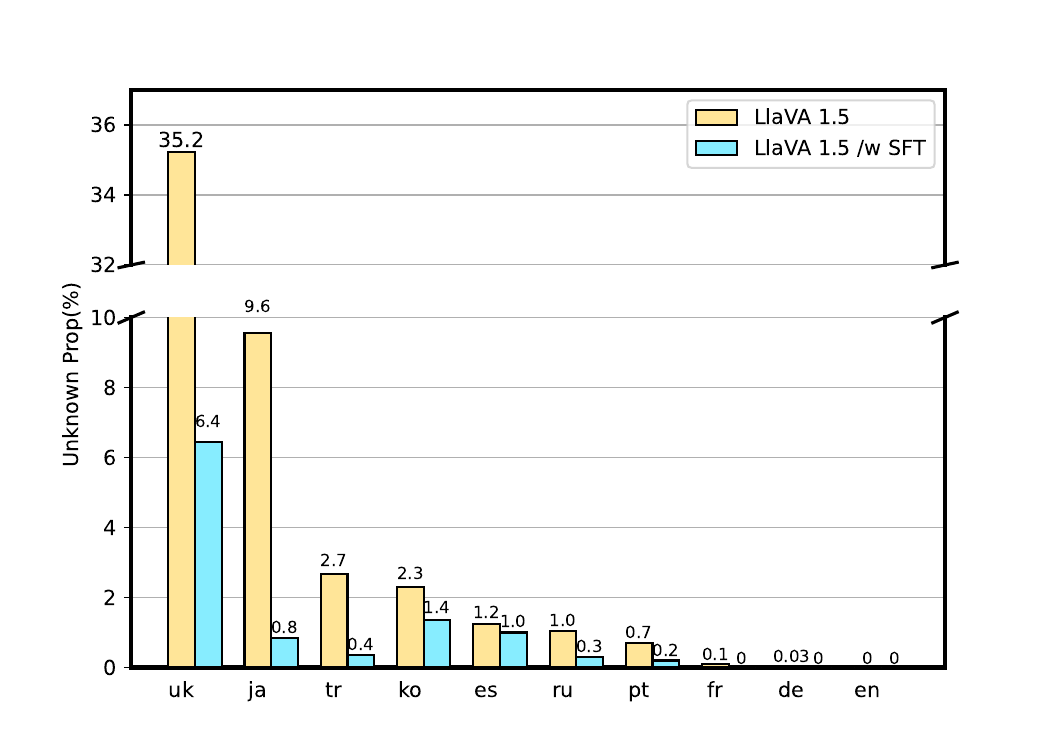}
\caption{``unknown prop'' refers to the ratio of invalid answers generated. After multilingual supervised instruction, the unknown prop of all presented languages significantly decreased.} 
\label{fig:instruction_following}
\end{figure}

\section{Method}

In this section, we present our two-stage Multilingual Hallucination Removal (MHR) framework for LVLMs. 
Given that hallucination refers to generating
output that is irrelevant or factually inaccurate given the input image-query pair, our two-stage framework focuses on reducing those irrelevant or factually inaccurate outputs. 
Considering that most languages can not understand the input query accurately, leading to irrelevant responses and resulting in severe multilingual hallucinations. 
Our MHR first performs multilingual supervised fine-tuning to augment the instruction-following ability for all languages. 
Subsequently, we aim to improve the model’s ability to resist hallucinations. In this stage, we construct hallucination-aware data pairs for direct preference optimization. 
To generate abundant training data with rich diversity, based on an English hallucination-aware dataset, for each non-English language, we first use the LVLM to generate multiple responses for each query, and then identify the hallucination-aware pairs with a cross-lingual alignment. 
In this way, we make full use of the inherent capabilities of the LVLM
without manually gathering multilingual data. 
In conclusion, with this two-stage training, our method can achieve significant improvement in alleviating hallucinations.

\subsection{Multilingual Supervised Fine-tuning}

Given that most high-source and low-resource languages fail to follow the instructions of LVLMs, we can elucidate using a basic instance from the POPE benchmark. 
Given an input image, 
we pose a yes/no question to the LVLM model using non-English languages. However, the model often fails to generate an appropriate "yes" or "no" response. For instance, it might deliver responses like "YesNo" directly or produce irrelevant answers that bear no connection to the initial question.
This occurrence might be attributable to the existing training procedure of LVLMs, in which only English instruction data is employed.
As shown in Figure \ref{fig:instruction_following}, it is surprising to observe that both low-resource languages such as Ukrainian (uk) and high-resource languages like Japanese (ja) grapple with significant instruction-following issues in LVLM. 
In detail, uk has more than 35\% answers that can not follow instructions.
Consequently, there is an urgent need to improve the instruction-following ability in both low-resource and high-resource languages. 
To accomplish it, we finetune the LVLM with a multilingual instruction-following dataset PALO \cite{palo} ($D^h$) for supervised fine-tuning (SFT). 
In this stage, a token-level cross-entropy is employed to train our model LVLM-SFT $\pi^{SFT}$. 
During the SFT process, given an input query, the LVLM is forced to learn the format of provided answers in the dataset $D^h$. 
Thus, the model can learn to understand the intent of the input query. As illustrated in Figure \ref{fig:instruction_following}, with multilingual SFT, the rate of invalid answers significantly decreases. 
This reduction is particularly significant in the Japanese (ja) context, as the rate falls sharply from 9.6\% to 0.8\%. 
However, for non-English languages, there still exist instances of invalid responses. This phenomenon may be largely due to the scarcity of training corpora (e.g. UK) during the initial training phase of LLaMA. 
Hence, fully addressing the instruction-following problem in multilingual LVLM is an important area for future exploration.

\subsection{Construct Hallucination-Aware Data}
\label{sec:method_alignment}

In this stage, we focus on reducing factually inaccurate outputs by enhancing the model's resistance to hallucinations. To this end, we form hallucination-aware data pairs for direct preference optimization training. 
However, there is no available multilingual hallucination dataset for LVLMs. 
Instead of manual construction, 
a simple way is to translate the English hallucination-aware dataset into multiple languages. 
It can generate as much translation data for each language as the English dataset.
However, this method can only generate a limited number of training data and the response in the dataset lacks diversity. 
In this section, we propose harnessing the inherent capabilities of the LVLM by first generating a variety of multilingual responses and then selecting the proper hallucination-aware pairs through a cross-lingual alignment.

The following section will describe the detailed data construction process.

\subsubsection{Generating Non-English Responses} 

As shown in Figure \ref{SEN}, based on the supervised fine-tuned model LVLM-SFT $\pi^{SFT}$, and an English hallucination dataset \cite{zhao2023hallucinations} $D^h$ containing high-quality English positive-negative hallucination-aware data pairs, given a input image $v$, for each input query in $D^h$, 
we generate a variety of responses $E_{gen}=\{Y^{i}\}_{i=1}^N$ for each non-English languages with corresponding language input $q^{lang}$, where $N$ is a hyper-parameter representing the number of generated samples. 

\begin{equation}
Y^{i}=\{y_t|y_t\sim p_{\pi^{SFT}}(v,q^{lang},y_{<t})\} 
\end{equation}

In this paper, we generate 20 answers for each language to maximize the output diversity of the model.

\subsubsection{Cross-lingual Alignment}  
\label{sec:semantic_distance_estimate}

After generating non-English responses, it is necessary to arrange these responses and then choose the appropriate data pairs from them.
Thus, we align them with the English data pairs with a cross-lingual alignment. 
Specifically, we first translate the responses from other languages to English with an off-the-shelf translation model \cite{costa2022no} and then calculate the semantic distance between the translated responses and the original English answer in $D^h$. 
Here we propose two methods for computing semantic distance:

\begin{enumerate}

    \item \textbf{Scoring by Loss.}

    Given that the majority of machine translation models are optimized by calculating the cross-entropy loss, 
    we can measure the semantic distance by 
    the corresponding loss formulated as Equation \ref{eq_tf}:

    \begin{equation}
    \label{eq_tf}
    \begin{aligned}
    \mathcal{L}_{CE} = - \sum_{t=1}^{n} \log p_\theta(y_t|y_{<t},x)
    \end{aligned}
    \end{equation}

    where $x$ represents the input tokens from the translated response, $y$ denotes the token from the initial English response. 
    A larger loss value is obtained when the difference between the two sequences is more pronounced. 

    \item \textbf{Scoring by BLEU.}
    Metric BLEU \cite{papineni2002bleu} has been developed to gauge the translation quality. 
    It measures the extent to which machine-generated translations align with reference standards, with 
    higher scores indicating a stronger level of alignment between the two languages.

\end{enumerate}

With these above methods to estimate semantic distance, 
we can quantify the semantic distance score of each generated response. 
Thus, we can build the hallucination-aware data pairs from them according to the score. 
In the main experiments, we use the first method ``scoring by loss". However, scoring by BLEU can achieve better results in some languages. We compare them in Section \ref{sec:5.3}.

\subsubsection{Construct Explicit Hallucination-aware Pairs}  
\label{sec:3.2.3}
In this section, as the English hallucination dataset $D^h$ already contains a pair of hallucination response $Y_h$ and non-hallucination response $Y_{nh}$, we can construct explicit hallucination-aware pairs by aligning the non-English response $Y^{i}$ with them, namely the most similar responses to English hallucination and non-halluciantion answer will be chosen.   

Formally, this step first produces the distance scores $d_h$ and $d_{nh}$ as shown in below Equation \ref{eq2}:

\begin{equation}
\label{eq2}
\begin{aligned}
d_h= P(Y^{i};{Y}_{h};\theta_{trans}),  
d_{nh} = P(Y^{i};{Y}_{nh};\theta_{trans})
\end{aligned}
\end{equation}

Here $\theta_{trans}$ represents the offline machine translation model. 
$P(\cdot)$ denotes the calculation process of semantic distance. 
Subsequently, we rank the generated answers $E_{gen}$ based on the aforementioned scores $D_{h}=\{d_h\}_{i=1}^N$ and $D_{nh}=\{d_{nh}\}_{i=1}^N$, 
both of which are sorted in ascending order as small entropy loss indicates better alignment. 
Finally, we select the Top-$K$ items from rank lists $D_{nh}$ and $D_{h}$, thereby constructing the positive and negative sample pools for subsequent preference optimization. 
$K$ is a hyperparameter which controls the amount of data generated and we will analyze it in Section \ref{sec:5.4}. 
As the direct preference optimization requires a pair of positive and negative data, thus we can obtain $K^2$ pairs for each query, which significantly exceeds the amount of data obtained from translation and increases the diversity. 
Notably, the quantity generated can be controlled in a more flexible way. 

Finally, we repeat the above procedure for each language to get the final datasets for constructing explicitly hallucination-aware data pairs $D_e^h$ covering all non-English languages.

\subsubsection{Construct Implicit Hallucination-aware Pairs}

Considering that the generated non-English response may include both hallucination and non-hallucination responses, in this section, we propose to construct implicit hallucination-aware pairs by aligning the non-English response with the English non-hallucination answer $Y_{nh}$. In this way, the response with the greatest semantic distance from the English response might be the one with hallucinations. Furthremore, when constructing explicit hallucination-aware pairs, we construct the data according to the semantic distance, which could inadvertently prompt the model to correlate specific language patterns with hallucinations.
Thus, the implicit hallucination-aware pairs can also alleviate the model overfitting to a specific pattern.

In this section, we only align generated answers $E_{gen}=\{Y^{i}\}_{i=1}^n$ with English non-hallucination answer $Y_{nh}$. The semantic distance is computed and sorted as above $D_{nh}$. 
Subsequently, the Top-$K$ items are selected from  $D_{nh}$ as the positive sample pools, while the last Top-$K$ samples are designated as the negative sample pools. Similarly, we can also obtain $K^2$ postive-negative pairs for following DPO training. Finally, we can construct an implicitly hallucination-aware data pairs $D_i^h$ covering all non-English languages.

\subsection{Multilingual Direct Preference Optimization}

Through the procedure outlined in Section \ref{sec:method_alignment}, we can acquire positive and negative pairs for both explicit and implicit hallucination-aware datasets. Building upon the direct preference optimization (DPO), we further train the SFT LVLM $\pi_{SFT}$:

\begin{equation}
\label{eq4}
\scalebox{0.8}{$
\begin{aligned}
\mathcal{L}_{DPO}(\pi_\theta; \pi_{\text{ref}}) = -\mathbb{E}_{(q^{lang},v, Y_{pos}, Y_{neg}) \sim D}  \\
\left[ \log \sigma \left( \beta \log \frac{\pi_\theta(Y_{pos} |[v,q^{lang}])}{\pi_{\text{ref}}(Y_{pos} |[v,q^{lang}])} - \beta \log \frac{\pi_\theta(Y_{neg} |[v,q^{lang}])}{\pi_{\text{ref}}(Y_{neg} | [v,q^{lang}])} \right) \right]
\end{aligned}$}
\end{equation}

where $v$ and $q^{lang}$ stands for visual input and query of source language $lang$, $[\cdot]$ indicates the concatenation of vision and query features for LVLM and $\beta$ is a hyper-parameter. 
The reference and policy models are denoted by $\pi_{ref}$ and $\pi_{\theta}$, respectively. Both models, $\pi_{\theta}$ and $\pi_{ref}$, are initialized using the supervised fine-tuned model $\pi_{SFT}$. The term $\log \sigma$ refers to the log-sigmoid function. 

To conduct the optimization process, 
considering the explicit hallucination-aware dataset $D_e^h$ and implicit hallucination-aware dataset $D_i^h$ only contain non-English languages, we additionally translate the initial English dataset $D^h$ to multilingual $D_t^h$ to enrich the training process.
Thus, the dataset $D$ in this Equation contains data from $D_e^h$, $D_i^h$, and $D_t^h$. 
After this stage, the LVLM model favors non-halluciantion, thus effectively mitigating hallucinations.


\renewcommand{\arraystretch}{1.11}  
\begin{table*}[tp]
\centering
\caption{
Performances comparison of Base LLaVA-1.5, MBLIP, and our model on POPE benchmark's all three datasets. 
We evaluate on the ``popular" set. The average scores are marked in \colorbox{gray!20}{gray} and bold text denotes the best accuracy results.}
\label{tab:main_result_pope}
\resizebox{18cm}{!}{

\begin{tabular}{cc ||ccc|c || ccc|c || ccc|c}

\hline
\multicolumn{2}{c||}{} & \multicolumn{4}{c||}{MS COCO} & \multicolumn{4}{c||}{AOKVQA} & \multicolumn{4}{c}{GQA}\\
\textbf{lang.}& {method.} &Precision & Recall & F1 & Accuracy & Precision & Recall & F1 & Accuracy & Precision & Recall & F1 & Accuracy\\\hline

\multicolumn{14}{c}{\textbf{High Resource Languages (HRL)}} \\   \hline 
 \multirow{3}{*}{\centering \textbf{en}} 
 & Base & 94.6 & 78.1 & 85.6 & \textbf{86.0} & 87.9 & 87.0 & 87.4 & \textbf{86.8} & 83.0 & 84.0 & 83.5 & \textbf{82.8} \\ 
& MBLIP & 85.8 & 82.9 & 84.3 & 84.6 & 79.2 & 82.7 & 80.9 & 80.5 & 72.9 & 82.5 & 77.4 & 75.9 \\ 
& Ours & 97.3 & 70.7 & 81.9 & 84.4 & 91.8 & 77.3 & 83.9 & 85.2 & 87.6 & 76.1 & 81.4 & 82.6 \\ \hline 
 \multirow{3}{*}{\centering \textbf{ru}} 
 & Base & 53.6 & 98.4 & 69.4 & 55.9 & 52.1 & 99.9 & 68.5 & 53.9 & 52.6 & 99.7 & 68.9 & 54.8 \\ 
& MBLIP & 80.4 & 83.2 & 81.8 & 81.4 & 69.3 & 84.5 & 76.1 & \textbf{73.5} & 64.3 & 81.6 & 71.9 & \textbf{68.2} \\ 
& Ours & 80.7 & 91.5 & 85.8 & \textbf{84.5} & 64.8 & 96.6 & 77.6 & 72.0 & 58.7 & 97.3 & 73.2 & 64.3 \\ \hline 
 \multirow{3}{*}{\centering \textbf{de}} 
 & Base & 95.2 & 75.0 & 83.9 & 83.3 & 88.6 & 73.3 & 80.2 & 80.4 & 80.4 & 72.8 & 76.4 & 75.8 \\ 
& MBLIP & 50.0 & 99.9 & 66.7 & 50.0 & 50.0 & 100.0 & 66.7 & 50.0 & 50.0 & 100.0 & 66.7 & 50.0 \\ 
& Ours & 96.6 & 70.5 & 81.5 & \textbf{84.0} & 88.8 & 70.5 & 78.6 & \textbf{80.8} & 81.9 & 71.1 & 76.1 & \textbf{77.7} \\ \hline 
 \multirow{3}{*}{\centering \textbf{zh}} 
 & Base & 50.0 & 100.0 & 66.6 & 47.8 & 49.9 & 100.0 & 66.6 & 48.1 & 49.5 & 100.0 & 66.2 & 47.6 \\ 
& MBLIP & 87.4 & 72.5 & 79.2 & 81.0 & 79.1 & 75.5 & 77.3 & \textbf{77.8} & 72.6 & 72.2 & 72.4 & \textbf{72.5} \\ 
& Ours & 83.4 & 91.3 & 87.1 & \textbf{86.5} & 67.8 & 96.2 & 79.5 & 75.2 & 63.0 & 97.0 & 76.4 & 70.0 \\ \hline 
 \multirow{3}{*}{\centering \textbf{ja}} 
 & Base & 51.2 & 100.0 & 67.7 & 49.1 & 49.9 & 100.0 & 66.6 & 48.3 & 49.6 & 100.0 & 66.3 & 47.5 \\ 
& MBLIP & 50.6 & 99.8 & 67.2 & 51.3 & 50.0 & 99.8 & 66.7 & 50.1 & 50.1 & 99.7 & 66.7 & 50.1 \\ 
& Ours & 86.2 & 86.5 & 86.3 & \textbf{86.3} & 65.6 & 91.9 & 76.6 & \textbf{71.9} & 60.5 & 92.3 & 73.1 & \textbf{66.1} \\ \hline 
 \multirow{3}{*}{\centering \textbf{fr}} 
 & Base & 63.7 & 97.9 & 77.2 & 70.9 & 62.4 & 99.4 & 76.7 & 69.6 & 58.5 & 99.5 & 73.7 & 64.3 \\ 
& MBLIP & 92.2 & 29.3 & 44.4 & 58.9 & 90.6 & 34.3 & 49.7 & 61.7 & 83.6 & 30.1 & 44.3 & 59.2 \\ 
& Ours & 84.7 & 91.4 & 87.9 & \textbf{87.5} & 69.9 & 95.7 & 80.8 & \textbf{77.3} & 63.6 & 97.2 & 76.9 & \textbf{70.7} \\ \hline 
 \multirow{3}{*}{\centering \textbf{es}} 
 & Base & 97.7 & 47.7 & 64.1 & 73.2 & 95.9 & 48.3 & 64.2 & 73.0 & 95.8 & 47.3 & 63.3 & 72.6 \\ 
& MBLIP & 78.6 & 85.4 & 81.9 & 81.1 & 66.6 & 89.5 & 76.4 & 72.3 & 62.3 & 89.7 & 73.5 & 67.7 \\ 
& Ours & 94.9 & 76.5 & 84.7 & \textbf{85.6} & 82.7 & 83.9 & 83.3 & \textbf{83.1} & 75.4 & 84.4 & 79.6 & \textbf{78.2} \\ \hline 
 \multirow{3}{*}{\centering \textbf{pt}} 
 & Base & 64.4 & 97.8 & 77.7 & 70.8 & 60.5 & 99.5 & 75.2 & 66.6 & 57.5 & 99.7 & 73.0 & 62.6 \\ 
& MBLIP & 82.4 & 82.5 & 82.5 & 82.5 & 78.2 & 84.1 & 81.0 & \textbf{80.3} & 72.2 & 84.5 & 77.9 & \textbf{76.0} \\ 
& Ours & 80.8 & 91.0 & 85.6 & \textbf{84.7} & 67.5 & 95.3 & 79.0 & 74.7 & 61.2 & 96.2 & 74.8 & 67.6 \\ \hline 
\multirow{3}{*}{\centering \textbf{Avg.}} 
&\cellcolor{gray!20}  Base &\cellcolor{gray!20} 71.3 &\cellcolor{gray!20} 86.9 &\cellcolor{gray!20} 74.0 &\cellcolor{gray!20} 67.1 &\cellcolor{gray!20} 68.4 &\cellcolor{gray!20} 88.4 &\cellcolor{gray!20} 73.2 &\cellcolor{gray!20} 65.8 &\cellcolor{gray!20} 65.9 &\cellcolor{gray!20} 87.9 &\cellcolor{gray!20} 71.4 &\cellcolor{gray!20} 63.5 \\ 
&\cellcolor{gray!20}  MBLIP &\cellcolor{gray!20} 75.9 &\cellcolor{gray!20} 79.4 &\cellcolor{gray!20} 73.5 &\cellcolor{gray!20} 71.3 &\cellcolor{gray!20} 70.4 &\cellcolor{gray!20} 81.3 &\cellcolor{gray!20} 71.8 &\cellcolor{gray!20} 68.3 &\cellcolor{gray!20} 66.0 &\cellcolor{gray!20} 80.0 &\cellcolor{gray!20} 68.8 &\cellcolor{gray!20} 64.9 \\ 
&\cellcolor{gray!20}  Ours &\cellcolor{gray!20} 88.1 &\cellcolor{gray!20} 83.7 &\cellcolor{gray!20} 85.1 &\cellcolor{gray!20} \textbf{85.4} &\cellcolor{gray!20} 74.8 &\cellcolor{gray!20} 88.4 &\cellcolor{gray!20} 79.9 &\cellcolor{gray!20} \textbf{77.5} &\cellcolor{gray!20} 69.0 &\cellcolor{gray!20} 89.0 &\cellcolor{gray!20} 76.4 &\cellcolor{gray!20} \textbf{72.2} \\ \hline 
\multicolumn{14}{c}{\textbf{Low Resource Languages (LRL)}} \\  \hline 
 \multirow{3}{*}{\centering \textbf{uk}} 
 & Base & 41.6 & 100.0 & 58.8 & 16.8 & 49.7 & 100.0 & 66.4 & 18.2 & 52.3 & 100.0 & 68.7 & 18.3 \\ 
& MBLIP & 75.9 & 82.5 & 79.1 & \textbf{77.9} & 67.0 & 86.3 & 75.5 & \textbf{71.9} & 62.9 & 86.3 & 72.8 & \textbf{67.7} \\ 
& Ours & 50.4 & 98.7 & 66.8 & 49.3 & 49.5 & 99.5 & 66.1 & 48.4 & 49.6 & 99.4 & 66.2 & 48.1 \\ \hline 
 \multirow{3}{*}{\centering \textbf{bg}} 
 & Base & 49.8 & 100.0 & 66.5 & 48.7 & 49.3 & 100.0 & 66.0 & 48.5 & 49.3 & 100.0 & 66.0 & 48.5 \\ 
& MBLIP & 62.7 & 94.3 & 75.3 & \textbf{69.1} & 54.9 & 96.7 & 70.0 & 58.6 & 53.6 & 97.8 & 69.2 & 56.5 \\ 
& Ours & 63.2 & 92.8 & 75.2 & 68.5 & 61.9 & 95.5 & 75.1 & \textbf{66.2} & 58.9 & 97.4 & 73.4 & \textbf{61.4} \\ \hline 
 \multirow{3}{*}{\centering \textbf{tr}} 
 & Base & 0.0 & 0.0 & 0.0 & 49.7 & 0.0 & 0.0 & 0.0 & 49.9 & 0.0 & 0.0 & 0.0 & 49.7 \\ 
& MBLIP & 50.1 & 99.8 & 66.7 & 50.2 & 50.0 & 100.0 & 66.7 & 50.0 & 50.0 & 100.0 & 66.7 & 50.0 \\ 
& Ours & 89.3 & 50.1 & 64.2 & \textbf{71.6} & 82.5 & 53.9 & 65.2 & \textbf{70.3} & 79.3 & 50.4 & 61.6 & \textbf{67.7} \\ \hline 
 \multirow{3}{*}{\centering \textbf{ar}} 
 & Base & 64.3 & 93.8 & 76.3 & 68.3 & 58.6 & 94.2 & 72.2 & 62.3 & 57.9 & 96.0 & 72.2 & 61.7 \\ 
& MBLIP & 88.7 & 67.5 & 76.6 & 79.4 & 81.5 & 67.4 & 73.8 & 76.1 & 70.9 & 64.9 & 67.7 & 69.1 \\ 
& Ours & 88.4 & 86.0 & 87.2 & \textbf{87.3} & 73.1 & 89.7 & 80.6 & \textbf{78.4} & 65.1 & 92.6 & 76.5 & \textbf{71.5} \\ \hline 
 \multirow{3}{*}{\centering \textbf{ko}} 
 & Base & 65.2 & 62.0 & 63.5 & 44.7 & 49.8 & 59.8 & 54.4 & 37.2 & 54.5 & 60.4 & 57.3 & 37.5 \\ 
& MBLIP & 89.5 & 100.0 & 94.4 & 33.4 & 85.2 & 100.0 & 92.0 & 34.0 & 75.1 & 100.0 & 85.8 & 31.1 \\ 
& Ours & 80.6 & 67.5 & 73.4 & \textbf{72.0} & 76.7 & 68.2 & 72.2 & \textbf{67.4} & 75.0 & 68.0 & 71.3 & \textbf{66.0} \\ \hline 
\multirow{3}{*}{\centering \textbf{Avg.}} 
&\cellcolor{gray!20} Base &\cellcolor{gray!20} 44.2 &\cellcolor{gray!20} 71.2 &\cellcolor{gray!20} 53.0 &\cellcolor{gray!20} 45.7 &\cellcolor{gray!20} 41.5 &\cellcolor{gray!20} 70.8 &\cellcolor{gray!20} 51.8 &\cellcolor{gray!20} 43.2 &\cellcolor{gray!20} 42.8 &\cellcolor{gray!20} 71.3 &\cellcolor{gray!20} 52.8 &\cellcolor{gray!20} 43.1 \\ 
&\cellcolor{gray!20}  MBLIP &\cellcolor{gray!20} 73.4 &\cellcolor{gray!20} 88.8 &\cellcolor{gray!20} 78.4 &\cellcolor{gray!20} 62.0 &\cellcolor{gray!20} 67.7 &\cellcolor{gray!20} 90.1 &\cellcolor{gray!20} 75.6 &\cellcolor{gray!20} 58.1 &\cellcolor{gray!20} 62.5 &\cellcolor{gray!20} 89.8 &\cellcolor{gray!20} 72.4 &\cellcolor{gray!20} 54.9 \\ 
&\cellcolor{gray!20}  Ours &\cellcolor{gray!20} 74.4 &\cellcolor{gray!20} 79.0 &\cellcolor{gray!20} 73.4 &\cellcolor{gray!20} \textbf{69.7} &\cellcolor{gray!20} 68.7 &\cellcolor{gray!20} 81.4 &\cellcolor{gray!20} 71.8 &\cellcolor{gray!20} \textbf{66.1} &\cellcolor{gray!20} 65.6 &\cellcolor{gray!20} 81.5 &\cellcolor{gray!20} 69.8 &\cellcolor{gray!20} \textbf{62.9} \\ \hline \hline

\end{tabular}
}

\end{table*}
\renewcommand{\arraystretch}{1} 


\begin{table*}[ht]
\caption{Results on the hallucination subset of MME. Higher scores indicate better performance and fewer hallucinations. 
The best performances within each setting are \textbf{bolded}. Limited by space, we only present 4 languages here, including high-resource languages ru and zh, and low-resource languages uk and bg. To help understand the overall performance comparison, we also report the average results for all 13 languages. 
}
\centering
\resizebox{\linewidth}{!}{%

\begin{tabular}{@{}cc cccc cccc c@{}}
\toprule
\multirow{2}{*}{Language}& \multirow{2}{*}{Method} & \multicolumn{2}{c}{\textit{Existence}$\uparrow$} & \multicolumn{2}{c}{\textit{Count}$\uparrow$} & \multicolumn{2}{c}{\textit{Position}$\uparrow$} & \multicolumn{2}{c}{\textit{Color}$\uparrow$} & \multicolumn{1}{c}{\multirow{2}{*}{Total Scores$\uparrow$}} \\
&&  \textit{ACC} & \textit{ACC+}&\textit{ACC} & \textit{ACC+}&\textit{ACC} & \textit{ACC+}&\textit{ACC} & \textit{ACC+}& \\
\hline
\multirow{2}{*}{Russian (ru)}  & Base & $65.00$ & $30.00$ & $50.00$ & $0.00$ & $\textbf{50.00}$ & $\textbf{0.00}$ & $50.00$ & $0.00$  & $245.00$ \\ 
& Ours & $\textbf{86.67}$ & $\textbf{73.33}$ & $\textbf{63.33}$ & $\textbf{30.00}$ & $\textbf{50.00}$ & $\textbf{0.00}$ & $\textbf{70.00}$ & $\textbf{40.00}$  & $\textbf{413.33}$ \\ 
\hline 

\multirow{2}{*}{Chinese (zh)}  & Base & $68.33$ & $36.67$ & $50.00$ & $0.00$ & $50.00$ & $0.00$ & $53.33$ & $6.67$  & $265.00$ \\ 
& Ours & $\textbf{91.67}$ & $\textbf{83.33}$ & $\textbf{73.33}$ & $\textbf{46.67}$ & $\textbf{58.33}$ & $\textbf{16.67}$ & $\textbf{68.33}$ & $\textbf{36.67}$  & $\textbf{475.00}$ \\ 
\hline
\multirow{2}{*}{Ukrainian (uk)}  & Base & $65.00$ & $30.00$ & $63.33$ & $26.67$ & $50.00$ & $0.00$ & $55.00$ & $10.00$  & $300.00$ \\ 
& Ours & $\textbf{95.00}$ & $\textbf{90.00}$ & $\textbf{88.33}$ & $\textbf{76.67}$ & $\textbf{66.67}$ & $\textbf{33.33}$ & $\textbf{91.67}$ & $\textbf{83.33}$  & $\textbf{625.00}$ \\ 
\hline
\multirow{2}{*}{Bulgarian (bg)}  & Base & $56.67$ & $13.33$ & $50.00$ & $0.00$ & $50.00$ & $0.00$ & $50.00$ & $0.00$  & $220.00$ \\ 
& Ours & $\textbf{70.00}$ & $\textbf{43.33}$ & $\textbf{71.67}$ & $\textbf{46.67}$ & $\textbf{51.67}$ & $\textbf{3.33}$ & $\textbf{70.00}$ & $\textbf{40.00}$  & $\textbf{396.67}$ \\ 
\hline
\multirow{2}{*}{Average 13 languages.}  
&\cellcolor{gray!20} Base &\cellcolor{gray!20} $70.77$ &\cellcolor{gray!20} $45.64$ &\cellcolor{gray!20} $53.72$ &\cellcolor{gray!20} $14.87$ &\cellcolor{gray!20} $49.87$ &\cellcolor{gray!20} $8.72$ &\cellcolor{gray!20} $59.87$ &\cellcolor{gray!20} $24.36$  &\cellcolor{gray!20} $327.82$ \\ 
&\cellcolor{gray!20}  Ours &\cellcolor{gray!20} $\textbf{81.54}$ &\cellcolor{gray!20} $\textbf{69.49}$ &\cellcolor{gray!20} $\textbf{65.26}$ &\cellcolor{gray!20} $\textbf{42.31}$ &\cellcolor{gray!20} $\textbf{53.08}$ &\cellcolor{gray!20} $\textbf{14.87}$ &\cellcolor{gray!20} $\textbf{71.03}$ &\cellcolor{gray!20} $\textbf{49.23}$  &\cellcolor{gray!20} $\textbf{446.79}$ \\ 
\hline


\end{tabular}
}
\label{tab:mme}
\end{table*}

\begin{table}[t!]
\centering
\caption{
Comparison between the basic LLaVA 1.5 and the model with our MHR framework. 
Qualified Content (QC) reported here means the ratio of generated text in the correct language in all generated text. 
Avg$^\text{H}$, Avg$^\text{L}$, Avg$^\text{A}$ represent the average accuracy of high-resource and low-resource languages, and all languages respectively.}
\begin{tabular}{cc|ccccc}
\toprule
Language & Method & CHAIR$ \downarrow $ & Cover$ \uparrow $ & HAL$ \downarrow $ & QC$ \uparrow $ \\
\midrule
\multirow{2}{*}{\centering \textbf{en}}& Baseline & 6.1& 34.3& 13.2& \textbf{100.0}\\
& Ours & \textbf{5.0}& \textbf{34.5}& \textbf{11.1}& \textbf{100.0}\\\hline
\multirow{2}{*}{\centering \textbf{ru}}& Baseline & 99.5& 0.1& 99.6& 0.5\\
& Ours & \textbf{74.6}& \textbf{4.4}& \textbf{77.1}& \textbf{100.0}\\\hline
\multirow{2}{*}{\centering \textbf{de}}& Baseline & 99.8& 0.0& 99.8& 0.0\\
& Ours & \textbf{14.8}& \textbf{23.3}& \textbf{27.0}& \textbf{99.8}\\\hline
\multirow{2}{*}{\centering \textbf{zh}}& Baseline & 92.3& 1.4& 94.3& 9.7\\
& Ours & \textbf{13.1}& \textbf{23.5}& \textbf{23.8}& \textbf{99.9}\\\hline
\multirow{2}{*}{\centering \textbf{ja}}& Baseline & 33.0& 14.9& 46.8& 91.0\\
& Ours & \textbf{19.8}& \textbf{20.4}& \textbf{32.4}& \textbf{100.0}\\\hline
\multirow{2}{*}{\centering \textbf{fr}}& Baseline & 100.0& 0.0& 100.0& 0.0\\
& Ours & \textbf{11.4}& \textbf{30.5}& \textbf{24.3}& \textbf{100.0}\\\hline
\multirow{2}{*}{\centering \textbf{es}}& Baseline & 98.7& 0.2& 98.8& 0.1\\
& Ours & \textbf{11.4}& \textbf{29.0}& \textbf{21.9}& \textbf{99.9}\\\hline
\multirow{2}{*}{\centering \textbf{pt}}& Baseline & 92.1& 1.3& 95.4& 3.0\\
& Ours & \textbf{14.0}& \textbf{27.8}& \textbf{26.6}& \textbf{100.0}\\\hline
\multirow{2}{*}{\centering \textbf{Avg$^\text{H}$.}}& \cellcolor{gray!20}  Baseline & \cellcolor{gray!20}77.7& \cellcolor{gray!20}6.5& \cellcolor{gray!20}81.0& \cellcolor{gray!20}25.5\\
& \cellcolor{gray!20} Ours & \cellcolor{gray!20}\textbf{20.5}& \cellcolor{gray!20}\textbf{24.2}& \cellcolor{gray!20}\textbf{30.5}& \cellcolor{gray!20}\textbf{100.0}\\\hline
\multirow{2}{*}{\centering \textbf{uk}}& Baseline & 64.5& 6.4& 66.8& 70.3\\
& Ours & \textbf{53.9}& \textbf{8.4}& \textbf{58.4}& \textbf{99.9}\\\hline
\multirow{2}{*}{\centering \textbf{bg}}& Baseline & 99.9& 0.0& 100.0& 0.2\\
& Ours & \textbf{52.8}& \textbf{8.9}& \textbf{60.3}& \textbf{75.5}\\\hline
\multirow{2}{*}{\centering \textbf{tr}}& Baseline & 96.9& 0.5& 97.5& 7.0\\
& Ours & \textbf{60.6}& \textbf{5.0}& \textbf{78.5}& \textbf{58.1}\\\hline
\multirow{2}{*}{\centering \textbf{ar}}& Baseline & 42.7& 10.3& 70.7& 80.6\\
& Ours & \textbf{29.4}& \textbf{20.4}& \textbf{48.8}& \textbf{100.0}\\\hline
\multirow{2}{*}{\centering \textbf{ko}}& Baseline & 100.0& 0.0& 100.0& 0.6\\
& Ours & \textbf{48.9}& \textbf{6.9}& \textbf{67.7}& \textbf{100.0}\\\hline
\rowcolors{1}{gray!20}{white} 
\multirow{2}{*}{\centering \textbf{Avg$^\text{L}$.}} &   \cellcolor{gray!20}Baseline &\cellcolor{gray!20} 80.8& \cellcolor{gray!20}3.4& \cellcolor{gray!20}87.0& \cellcolor{gray!20}31.7\\
& \cellcolor{gray!20}  Ours & \cellcolor{gray!20}\textbf{49.1}& \cellcolor{gray!20}\textbf{9.9}& \cellcolor{gray!20}\textbf{62.7}& \cellcolor{gray!20}\textbf{86.7}\\\hline
\multirow{2}{*}{\centering \textbf{Avg$^\text{A}$.}}& \cellcolor{gray!20} Baseline &\cellcolor{gray!20} 78.9& \cellcolor{gray!20}5.3& \cellcolor{gray!20}83.3& \cellcolor{gray!20}27.9\\
& \cellcolor{gray!20} Ours & \cellcolor{gray!20}\textbf{31.5}& \cellcolor{gray!20}\textbf{18.7}& \cellcolor{gray!20}\textbf{42.9}& \cellcolor{gray!20}\textbf{94.9}\\\hline 

\end{tabular}
\label{tab:results}
\end{table}

\begin{figure*}[!tp]
  \centering
  \subfloat[uk]{\includegraphics[width=0.24\textwidth]{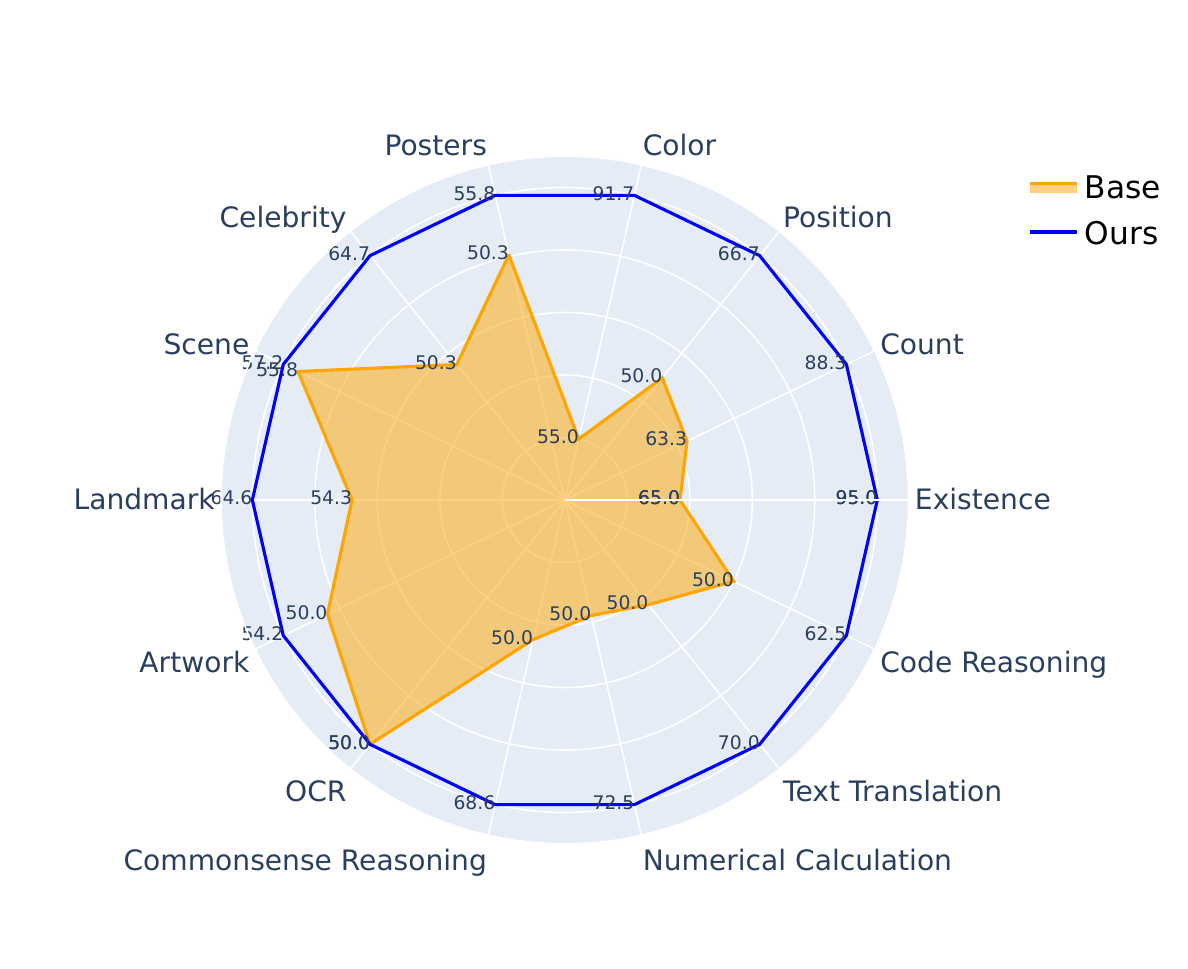}}
  \hfil
  \subfloat[zh]{\includegraphics[width=0.24\textwidth]{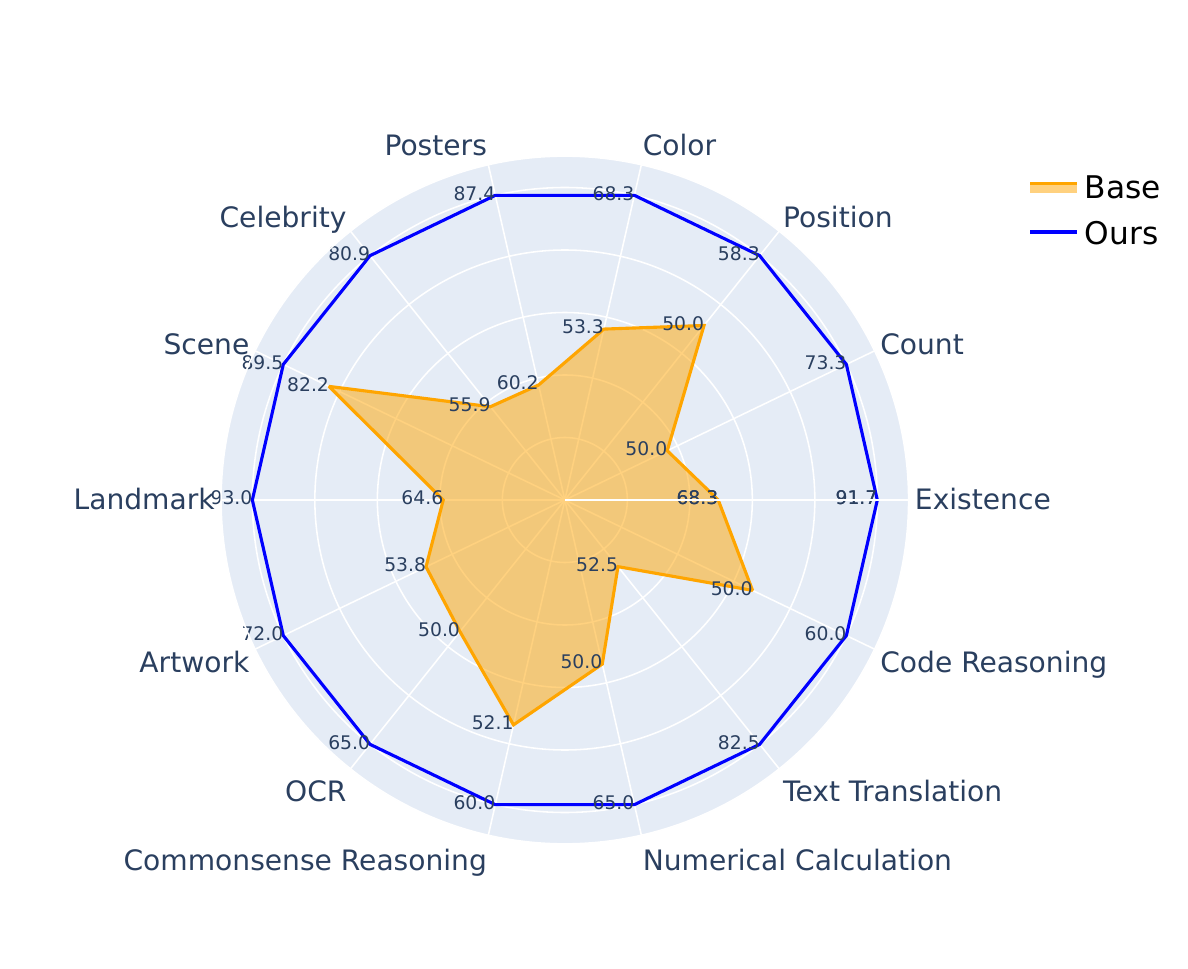}}
  \hfil
  \subfloat[bg]{\includegraphics[width=0.24\textwidth]{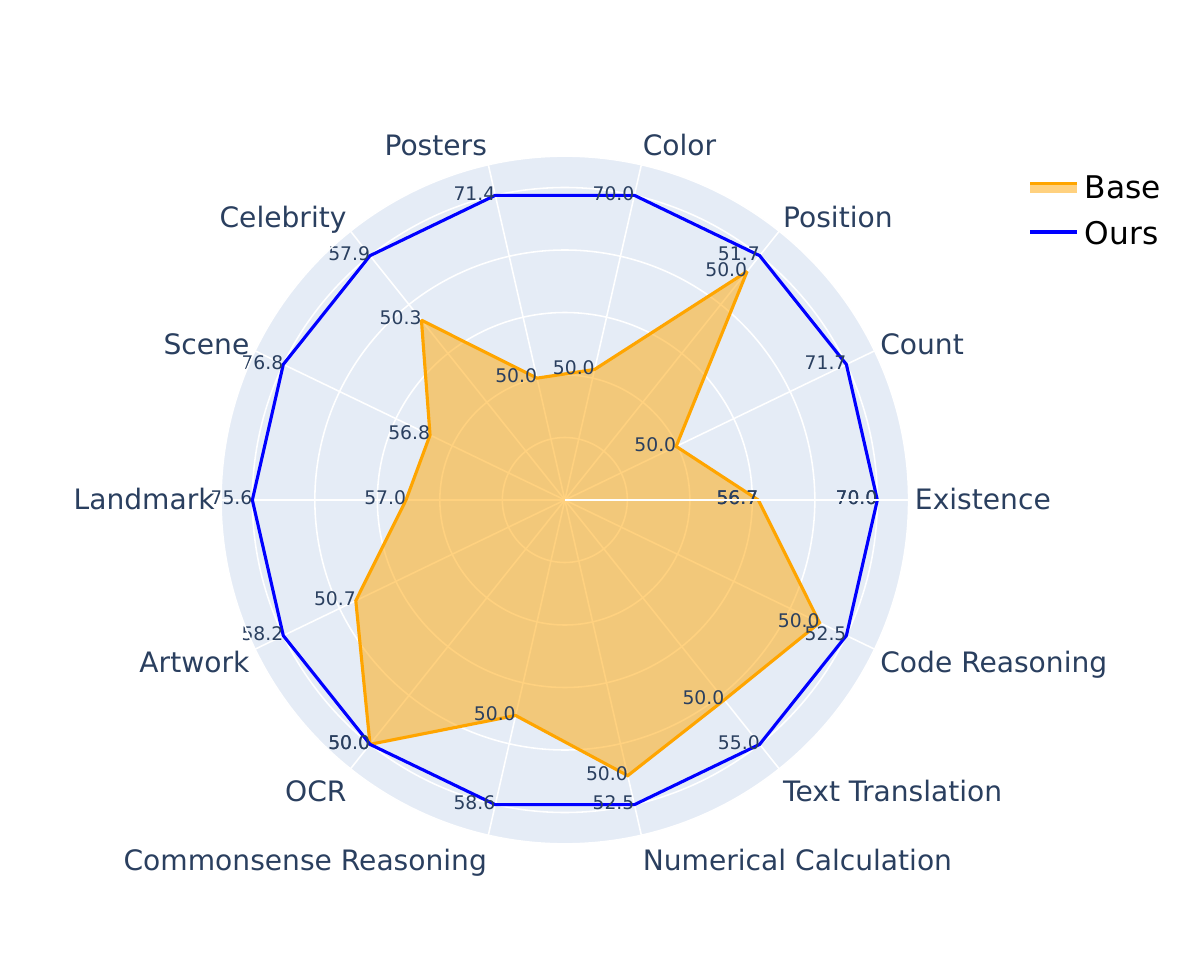}}
  \hfil
  \subfloat[ru]{\includegraphics[width=0.24\textwidth]{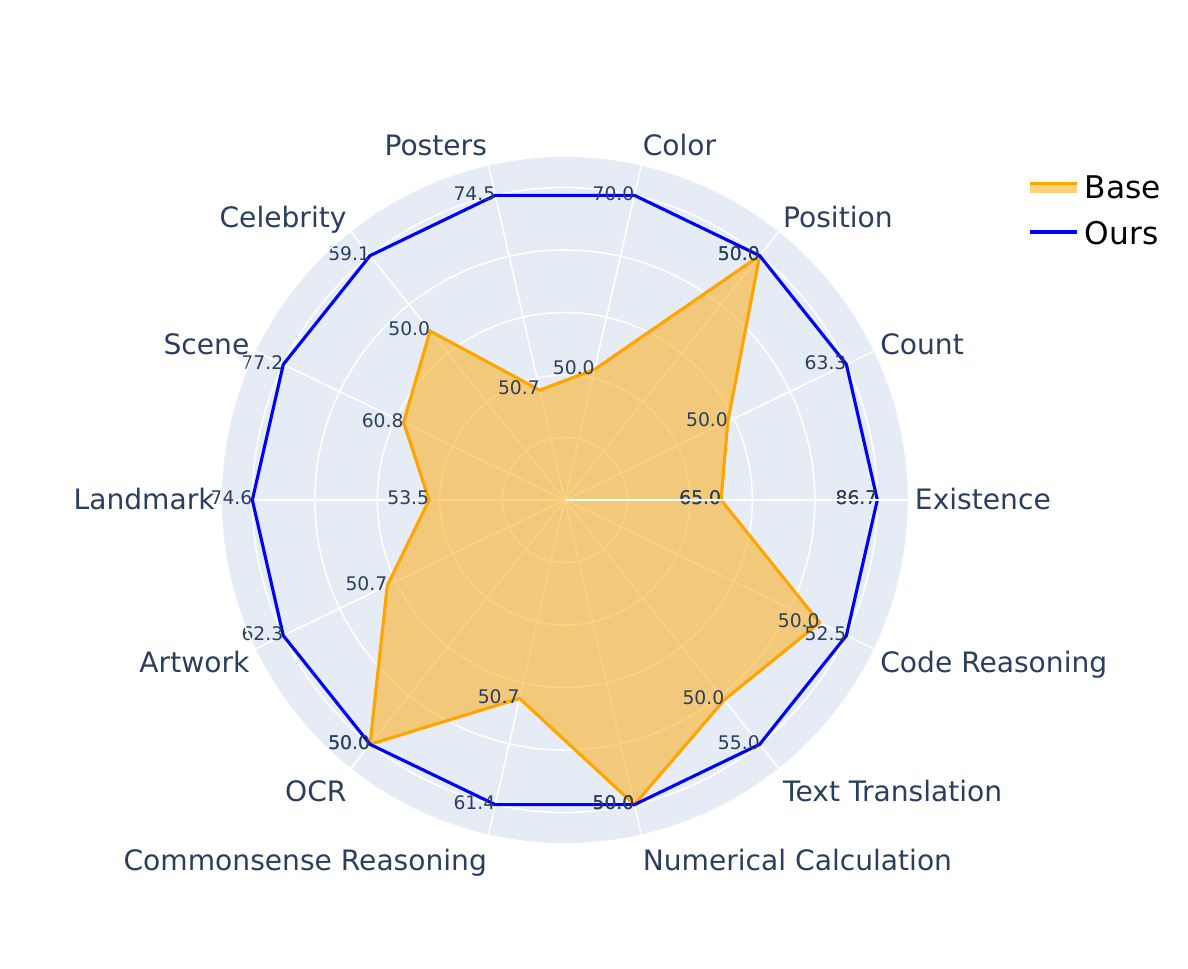}}
  \caption{The performance on the full MME set, which consists of 14 tasks. Each graph displays the performance of the respective LLaVA-1.5 and our MHR model. Here we present results in four languages (uk, zh, bg, and ru) as outlined in Table \ref{tab:mme}.}
  \label{fig:mme_full}

\end{figure*}
\section{Experiment}
 
\subsection{Evaluation Metrics}
Following previous works \cite{leng2023VCD,huang2023opera} mitigating hallucinations in LVLMs, in our experiments, we adopt two widely-used discriminative benchmarks POPE and MME. {In addition, to analyze the performance of our method on generative tasks, we further employ a generative benchmark AMBER for evaluation.} As our method is devised for mitigating multilingual hallucination, before evaluation, we first extend the above three benchmarks to the multilingual version. 

\textbf{POPE MUL} \cite{Li2023POPE}, 
the standard Polling-based Object Probing Evaluation (POPE) offers a comprehensive method for evaluating object hallucination. It queries LVLMs about the existence of a specific object in the given image in a balanced manner, with a 50:50 ratio for existent versus non-existent objects. It utilizes three unique sampling methods: random, popular, and adversarial, with differences in their processes of creating negative samples. The POPE benchmark integrates data from three distinct repositories, namely MSCOCO \cite{lin2014microsoft}, A-OKVQA \cite{schwenk2022okvqa}, and GQA \cite{hudson2019gqa}. For every sampling approach, it uses 500 images from each dataset, and develops six questions for each image, resulting in a total of 27,000 query-answer pairs from the development sets of these mentioned datasets. The evaluation is based on four vital metrics: Accuracy, Precision, Recall, and the F1 score. Significantly, in the computation of accuracy, the model dispensing irrelevant answers is flagged as incorrect. This circumstance is not so prevalent in English but is notably common in other languages.

As the language of annotation in the POPE benchmark is exclusively English, precluding its use for multilingual evaluations. Therefore, we introduce \textbf{POPE MUL}, an expansion of the POPE benchmark to a multilingual version.
Specifically, we first adopt GPT-3.5 to translate all queries and answers of POPE benchmarks into 12 different languages and meticulously curate the translation results to maintain superior benchmark quality.

\textbf{MME MUL}  MME \cite{fu2023MME} serves as an extensive benchmark tailored to assess LVLMs across multiple dimensions. It comprises ten perception-related subtasks and four cognition-focused ones. Similarly, we translate MME into 12 languages utilizing GPT-3.5 and correct the mistakes during translation manually. 
We calculate the ACC and ACC+ of each task and the total score for comparison.

\textbf{AMBER MUL} AMBER \cite{wang2023llm} is an LLM-free multi-dimensional benchmark for evaluating hallucinations. It can be used to evaluate both generative task and discriminative task including existence, attribute, and relation hallucination. In this paper, we adopt this benchmark to investigate the generative ability of our method. {Specifically, we adopt four metrics including CHAIR \cite{rohrbach2018object}, Cover, and Hal for evaluating hallucinations on the generative task. 
Here CHAIR metric \cite{rohrbach2018object} measures the frequency of hallucinatory objects appearing in the responses. Cover measures the object coverage of responses, and Hal represents the proportion of responses with hallucinations. 
} 
Similar to the above two benchmarks, we extend the English-centric AMBER to a multilingual version of AMBER MUL.  
Notably, considering this is a scenario of generating content in multiple languages, we further devise a new metric ``Qualified Content'' which indicates the ratio of generated sentences that align with the target language. We use Google's language detection \footnote{https://pypi.org/project/langdetect/} to assess whether the generated sentences satisfy the criteria. With these four metrics, we can comprehensively evaluate the generative ability of the model.

\subsection{Implementation Details}

In this paper, we choose the widely-used LLaVA-1.5 as the base model. To further verify the effectiveness of our method, we also conduct experiments on the strong LVLM model CogVLM \cite{wang2023cogvlm}. 
We consider English and 12 commonly used languages including Russian (ru), German (de), Chinese (zh), Japanese (ja), French (fr), Spanish (es), Portuguese (pt), Ukrainian (uk), Bulgarian (bg), Turkish (tr), Arabic (ar), and Korean (ko).
For multilingual supervised fine-tuning (SFT), we leverage the multilingual instruction-following dataset in PALO \cite{palo} for training, which encompasses 2.08M instruction-answer data pairs. 
We use a batch size of 8 and employ four NVIDIA A100 80GB for training 1K steps. We found that more training steps in the SFT stage lead to a performance drop. 
For training settings, we utilize AdamW \cite{Loshchilov2017adamw} optimizer and the learning rate is set at 2e-5.

Regarding the DPO stage, we construct datasets based on HA-DPO \cite{zhao2023hadpo} $D^h$ which contains 1735 English hallucination and non-hallucination data pairs. For each language, to construct explicit hallucination-aware pairs $D_e^h$ and implicit hallucination-aware pairs $D_i^h$, we set $K=3$ and obtain 15615 data pairs for each construction method. Thus, for 12 languages, we have constructed a total of 374760 training pairs.
In addition, we also translate the data pairs from HA-DPO to multilingual dataset $D_t^h$, resulting to an additional 1735 training pairs for non-English languages.  
Although $D_e^h$ and $D_i^h$ are far greater in quantity than $D_c^h$, we train our final model utilizing these three data with the same sampling ratio. In this way, we can use the same amount of data from these three datasets.
In this paper, we treat each non-English language equally and use the same amount of data for training.

In the DPO training process, we use eight NVIDIA A100 80GB with a batch size of eight for every device. Moreover, we embrace a learning rate set at 2e-6 and establish $\beta$ at a fixed constant of 0.1. 

During evaluation, all models use a greedy decoding with temperature 0 for fair comparison.

\subsection{Main Results}

\textbf{Results on POPE MUL.} 
The results on POPE benchmark under the MSCOCO, A-OKVQA, and GQA datasets are summarized in Table \ref{tab:main_result_pope}. 
Aside from a slight decline in English, we have achieved significant improvements over the base model in all non-English languages. 
The observed decline in English is due to the fact that the amount of data in other languages during the training process far exceeds that of English.
This situation has also occurred in multilingual training in LLM \cite{shaham2024multilingual}. {Notably, this problem can be alleviated by supplementing English data \cite{shaham2024multilingual}. However, in this paper, we experiment are based on a fixed English dataset and focus on performance improvements across multiple languages.
}

In a more detailed analysis, on high-resource languages, our model achieves an average improvement of 18.3 points on MSCOCO, while 20.2\% (46.8 vs 67.0) improvement is obtained on the low-resource languages. 
With our method, we surprisingly found that querying by some high-resource languages (e.g. zh, ja, fr) even surpasses the original English performance 86.0 on MSCOCO. This result demonstrates the effectiveness of our MHR framework. In addition, on more difficult question answering datasets AOKVQA and GQA, our model can still achieve robust enhancements. 

To further demonstrate the effectiveness of our method, we compare our model with a strong multilingual model MBLIP \cite{geigle2023mblip}, {which re-align the image encoder previously tuned to an English LLM to a new, multilingual LLM and yields very competitive results.}
As depicted in Table \ref{tab:main_result_pope}, our model achieves significantly better results on high-resource languages than MBLIP. For low-resource languages, MBLIP presents stronger performance in uk and bg while our model performs better on the other three languages. The comparison with MBLIP further presents the effectiveness of our framework.

\noindent\textbf{Results on MME MUL.} 
Here we first analyze the results of the hallucination subset of MME. This subset extends beyond POPE’s scope, encompassing both object-level and attribute-level hallucinations. 
In Table \ref{tab:mme}, we present the results of our model and base model LLaVA-1.5. It is obvious that our MHR method significantly improves the total scores for both high-resource and low-resource languages. 

To have an overall understanding of our method, we report the average performance of all 13 languages in the last row of Table \ref{tab:mme}. For total scores, our MHR framework increases by 118.97 over the base model (327.82 vs 446.79). 
Furthermore, we test our model on the full set of MME containing 14 tasks to verify the general ability of the model. As shown in Figure \ref{fig:mme_full}, all presented languages showcase improvement on both perception and cognition tasks.
These results further prove the effectiveness of our approach.

\noindent \textbf{Results on AMBER MUL. } {With the above two benchmarks, we have verified the effectiveness of our proposed MHR on discriminative tasks. In Table III, we demonstrate the performance comparison on generative tasks. It is obvious that our method significantly improves the ratio of qualified content.} In other words, after using our framework, the model's proportion of responses in the correct language has significantly increased. 
Notably, for high-resource languages, such as ru, de, fr, es, and pt, our method improves the
ratio to nearly 100\%. It is attributed to our two-stage training, especially the multilingual supervised fine-tuning. 
Moreover, for all 13 languages, our method decreases the average CHAIR from 78.9 to 31.5, achieving a significant improvement. This metric further demonstrates that our method effectively mitigates the hallucination in LVLMs. Besides QC and CHAIR metric, for other two metrics, namely Cover and HAL, our model also markedly improves the object coverage and reduces the proportion of hallucinations in the responses.

\begin{figure}

    \centering
    \includegraphics[width=0.88\linewidth]{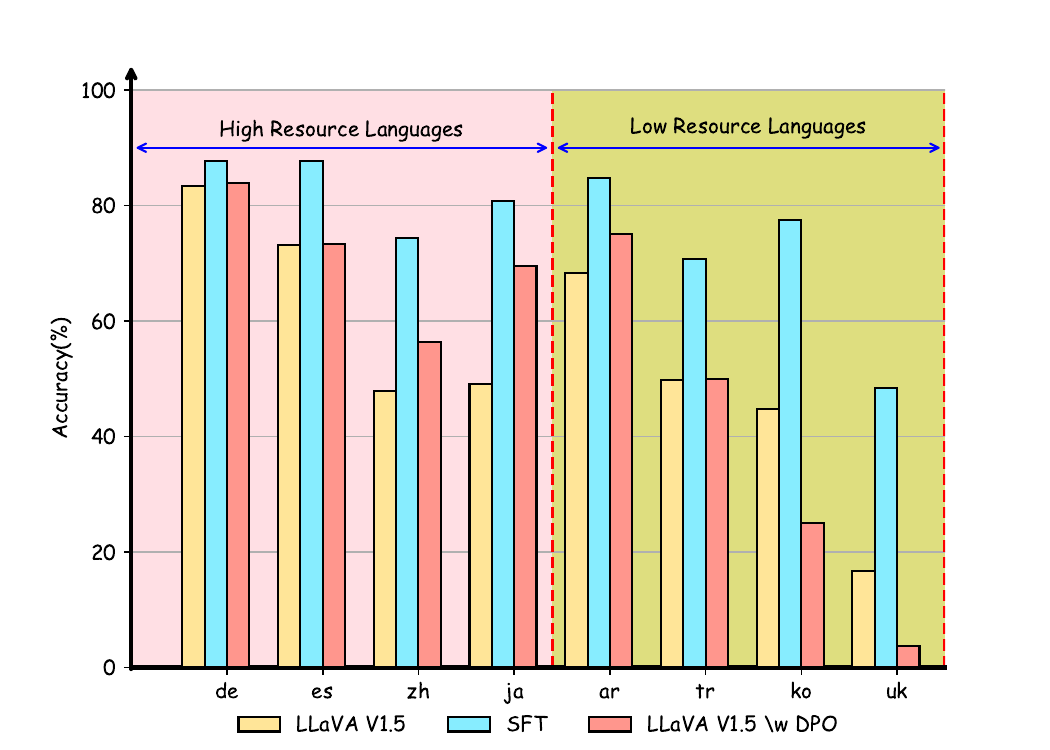}
    \caption{
    Comparison of base LLaVA 1.5, LLaVA 1.5 with SFT, and LLaVA 1.5 with direct DPO.
    }
    \label{fig:pre_dpo}
\end{figure}

\section{Ablation Study}

\subsection{Effectiveness of Supervised Fine-tuning.} 
 
In this section, we experiment to verify the effectiveness of supervised fine-tuning (SFT). According to Figure \ref{fig:pre_dpo}, when we initially apply SFT, we observe a rise in accuracy across all languages. Similar to Figure \ref{fig:instruction_following}, the reason may come from the decrease of non-sense answers. Meanwhile, the SFT can also boost the model's ability to accurately answer the input query. 
However, when we bypass the initial SFT stage and straightaway move to the DPO stage, the model only shows minimal improvements in certain languages and even exhibits performance degradation (e.g. ko, uk) in the remaining languages. Thus, SFT is essential for alleviating hallucinations in LVLMs for both following instructions and enhancing performance.

\subsection{Ablation of Different Datasets} 

In this paper, to obtain a superior multilingual model, we first perform SFT with a multilingual instruction following dataset $D_s$. Subsequently, we perform DPO with our constructed explicitly hallucination-aware dataset $D_e^h$, implicitly hallucination-aware dataset $D_i^h$, and translated multilingual dataset $D_t^h$. Thus, in this study, we conduct detailed ablation to analyze the effectiveness of each dataset. 
As shown in Table \ref{tab:model_performance_before}, we compute the average accuracy score on the POPE MSCOCO popular set. From this table, we can observe that: (1) Using our constructed hallucination-aware dataset $D_e^h$, $D_i^h$, or translated $D_t^h$ all bring improvement to the SFT LLaVA model. 
It verifies that our construct dataset helps alleviate multilingual hallucinations in LVLMs.
Meanwhile, the greatest improvement is achieved with translation data $D_t^h$, this is because this dataset is carefully constructed by humans.
(2) Training with both our construct data $D_e^h$ and $D_i^h$ leads to significant accuracy increases than using a single dataset, validating $D_i^h$ can help alleviate $D_e^h$'s overfitting to specific patterns and joint training on these two datasets contribute to better performance.  
(3) With all three datasets, we achieve the best average performance of 79.36. This result further confirms the effectiveness and necessity of our framework. Moreover, the dataset we built serves as a valuable complement to the translation dataset.

\newcommand{\smallcircle}{\tikz\draw (0,0) circle (.5ex);}

\begin{table}[t]
\caption{
Ablation study of different training data.  
In this study, we evaluate on POPE benchmark's popular set. We experiment on the LLaVA 1.5 model.}
\centering
\begin{tabular}{c|ccc|c}

\hline 

\textbf{Benchmark} & \textbf{$D_e^h$} & \textbf{$D_i^h$} & \textbf{$D_t^h$} & \textbf{Average Acc} \\ 
\hline 
\multirow{9}{*}{MS COCO} 

& \checkmark & \smallcircle  &\smallcircle & 76.58 \\
& \smallcircle &  \checkmark  & \smallcircle & 75.08 \\
& \smallcircle &  \smallcircle &  \checkmark& 77.11 \\
& \checkmark & \checkmark & \smallcircle & 78.00 \\
& \smallcircle  & \checkmark & \checkmark & 76.92 \\
& \checkmark & \smallcircle  & \checkmark & 79.12 \\
&\checkmark & \checkmark & \checkmark & \textbf{79.36} \\
&\multicolumn{3}{c|}{--- SFT LLaVA ---} & 74.61 \\
  &\multicolumn{3}{c|}{--- Base LLaVA ---} &  59.29 \\

\hline

\hline

\end{tabular}

\label{tab:model_performance_before}
\end{table}

\subsection{Semantic Distance Metric}

\label{sec:5.3}
In this section, we explore the impact of different metrics for semantic distance estimation as described in Section \ref{sec:semantic_distance_estimate}. The quality of the metric directly affects the quality of subsequent data construction.
As shown in Table \ref{tab:transferability}, we found that the Loss metric achieves better performance on high-resource ru (84.5 vs 76.8), fr (87.5 vs 81.30), and low-resource bg (68.5 vs 59.7) and the advantages of the Loss metric are quite obvious compared to BLUE metric. Thus, considering the overall performance, in this paper, we choose the Loss metric as the semantic distance metric.

\begin{table}[t]
\caption{
Ablation study on semantic distance estimation methods Loss and BLEU, as referred to in Section \ref{sec:semantic_distance_estimate}. 
}
\centering
\resizebox{0.92\linewidth}{!}{
\begin{tabular}{c cc|ccc|c}

\hline
\toprule
\textbf{Resource}  &\textbf{Language} &  \textbf{Method}          & \textbf{Precision} & \textbf{Recall}   & \textbf{F1}  &  \textbf{Acc.}   \\ \midrule

\multirow{6}{*}{HRL}

&\multirow{2}{*}{Russian (ru)}    & {Loss}&  80.7 & 91.5 & 85.8 & \textbf{84.5} \\
&&   {BLEU}   & 69.3 & 96.4 & 80.7 & 76.8 \\ \hhline{|~|-|-|-|-|-|-|} 
&\multirow{2}{*}{French (fr)}    & {Loss}& 84.7 & 91.4 & 87.9 & \textbf{87.5} \\
&&   {BLEU}   & 74.3 & 95.7 & 83.6 & 81.3  \\ \hhline{|~|-|-|-|-|-|-|} 
&\multirow{2}{*}{German (de)}    & {Loss}&  96.6 & 70.5 & 81.5 & 84.0 \\
&&  {BLEU}   & 93.6 & 79.7 & 86.1 & \textbf{87.1}\\ 
\hline

\multirow{6
}{*}{LRL} 
&\multirow{2}{*}{Arabic (ar)}   & {Loss}     & 88.4 & 86.0 & 87.2 & 87.3 \\
&&   {BLEU}     & 86.0 & 92.3 & 89.0 & \textbf{88.6} \\ \hhline{|~|-|-|-|-|-|-|}

&\multirow{2}{*}{Bulgarian (bg)}   & {Loss}     & 63.2 & 92.8 & 75.2 & \textbf{68.5} \\
&&   {BLEU}     & 55.5 & 97.5 & 70.8 & 59.7  \\ \hhline{|~|-|-|-|-|-|-|} 
              
&\multirow{2}{*}{Turkish (tr) }   & {Loss}    & 89.3 & 50.1 & 64.2 & 71.6 \\
&&   {BLEU}     &  89.3 & 56.9 & 69.5 & \textbf{74.5} \\ 

\hline
\bottomrule
\end{tabular}
}

\label{tab:transferability}
\end{table}

\begin{figure}

\centering
\includegraphics[width=8.5cm]{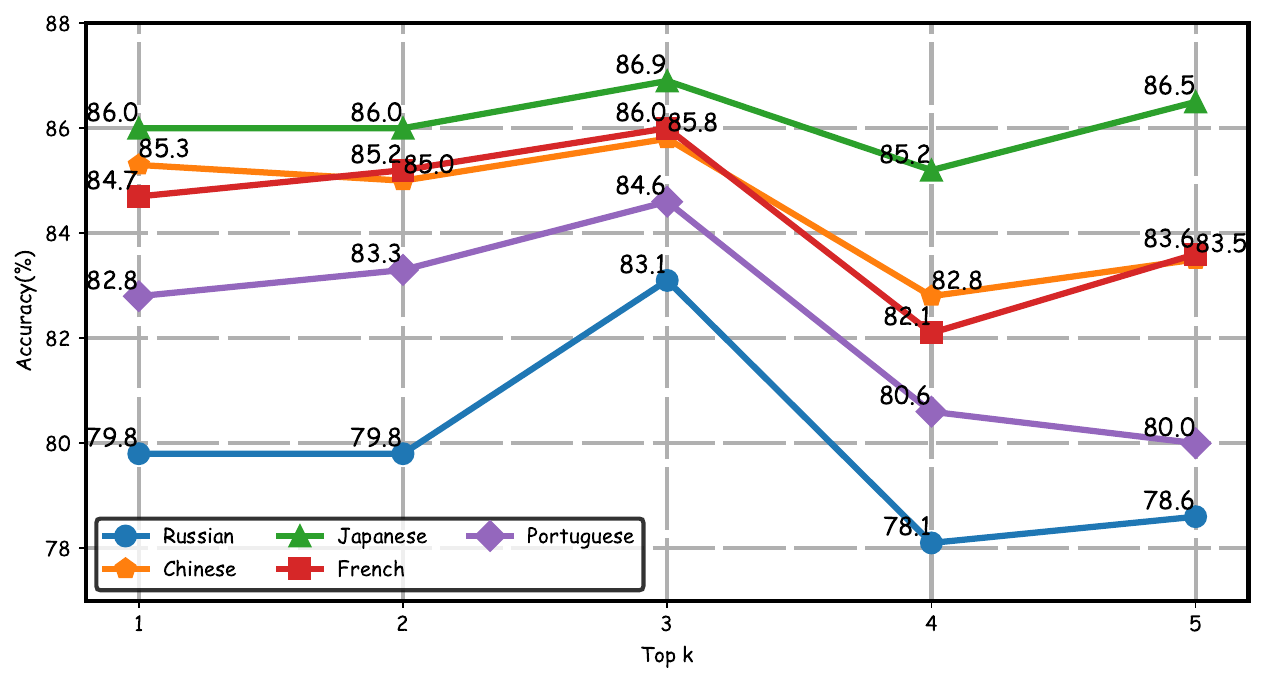}
\caption{The ablation results on top-$K$ in Section \ref{sec:3.2.3}. An increased value of K implies that more data will be generated.}  
\label{fig:topk_ablation}
\end{figure}

\subsection{DPO Training Data}
\label{sec:5.4}
In this section, we study the impact of the amount of generated hallucination-aware data $D_e^h$ and $D_i^h$ on the final model effect. Due to computational resource constraints, we trained each variant for 5.5K steps, 
assuring each model processed identical data volumes by modulating the sampling rate.
As shown in Figure \ref{fig:topk_ablation}, when K is set to 3, the best performance is achieved on all languages. It indicates that an optimal balance between the quantity and quality of construct data is realized at K equals 3. 
Therefore, our final model in Table \ref{tab:main_result_pope} is implemented with K equal to 3, and it has been run for an extended number of 24k training steps.

\subsection{Ablation of Different Resource Ratios}
{In this paper, we reveal the severe multilingual hallucination problem in LVLMs, covering both high-resource and low-resource languages. 
In our model, we propose to alleviate the hallucinations for all these languages, thus treating them equally, namely the data sampling ratio of high-resource languages and low-resource languages is 1:1. To study the impact of divergent ratios between high-resource and low-resource languages on performance, we perform an ablation study in this section. By training our model with differing data sampling ratios (i.e., high: low = 1:2), and evaluating them within the POPE MUL benchmark, we gain extensive insights. The outcomes of this exploration are detailed in Table \ref{tab:high_low_ratio}. Referring to this table, when increasing the sampling ratio of low-resource languages, some low-resource languages can achieve better results, such as ar and ko. However, the average accuracy of both high-resource and low-resource languages will decrease. Thus, it is more appropriate to treat these languages equally when considering both high-resource and low-resource languages.}


\begin{table}
\caption{
The ratio `x:y' denotes the sample rate for high-resource languages versus that of low-resource languages. 
Our evaluation is conducted on the ``popular'' dataset subset. Average scores are indicated in a \colorbox{gray!20}{gray} shade while the top accuracy results are highlighted in bold.
}
\label{tab:high_low_ratio}
\centering
\scalebox{0.98}{
\begin{tabular}{cc |ccc|c }

\hline
\textbf{Language}& {Method} &Precision & Recall & F1 Score& Acc. \\
\hline

\multicolumn{6}{c}{\textbf{High Resource Languages (HRL)}} \\   \hline 
 \multirow{2}{*}{\centering \textbf{en}} 
& 1:1 & 97.3 & 70.7 & 81.9 & 84.4 \\ 
& 1:2 & 97.1 & 72.0 & 82.7 & 84.9 \\ 
\hline 
 \multirow{2}{*}{\centering \textbf{ru}} 
& 1:1 & 80.7 & 91.5 & 85.8 & \textbf{84.5} \\ 
& 1:2 & 73.3 & 94.8 & 82.7 & 80.1 \\ 
\hline 
 \multirow{2}{*}{\centering \textbf{de}} 
& 1:1 & 96.6 & 70.5 & 81.5 & 84.0 \\ 
& 1:2 & 95.7 & 74.3 & 83.7 & \textbf{85.5} \\ 
\hline 
 \multirow{2}{*}{\centering \textbf{zh}} 
& 1:1 & 83.4 & 91.3 & 87.1 & \textbf{86.5} \\ 
& 1:2 & 76.0 & 95.0 & 84.4 & 82.5 \\ 
\hline 
 \multirow{2}{*}{\centering \textbf{ja}} 
& 1:1 & 86.2 & 86.5 & 86.3 & 86.3 \\ 
& 1:2 & 83.9 & 89.9 & 86.8 & \textbf{86.3} \\ 
\hline 
 \multirow{2}{*}{\centering \textbf{fr}} 
& 1:1 & 84.7 & 91.4 & 87.9 & \textbf{87.5} \\ 
& 1:2 & 79.6 & 93.7 & 86.1 & 84.8 \\ 
\hline 
 \multirow{2}{*}{\centering \textbf{es}} 
& 1:1 & 94.9 & 76.5 & 84.7 & 85.6 \\ 
& 1:2 & 93.3 & 79.7 & 86.0 & \textbf{86.9} \\ 
\hline 
 \multirow{2}{*}{\centering \textbf{pt}} 
& 1:1 & 80.8 & 91.0 & 85.6 & \textbf{84.7} \\ 
& 1:2 & 77.5 & 93.2 & 84.6 & 83.0 \\ 
\hline 
\multirow{2}{*}{\centering \textbf{Avg.}} 
&\cellcolor{gray!20}  1:1 &\cellcolor{gray!20} 88.1 &\cellcolor{gray!20} 83.7 &\cellcolor{gray!20} 85.1 &\cellcolor{gray!20} \textbf{85.4} \\ 
&\cellcolor{gray!20}  1:2 &\cellcolor{gray!20} 84.5 &\cellcolor{gray!20} 86.6 &\cellcolor{gray!20} 84.6 &\cellcolor{gray!20} 84.3 \\ 
\hline 
\multicolumn{6}{c}{\textbf{Low Resource Languages (LRL)}} \\  \hline 
 \multirow{2}{*}{\centering \textbf{uk}} 
& 1:1 & 50.4 & 98.7 & 66.8 & \textbf{49.3} \\ 
& 1:2 & 49.5 & 98.8 & 66.0 & 48.0 \\ 
\hline 
 \multirow{2}{*}{\centering \textbf{bg}} 
& 1:1 & 63.2 & 92.8 & 75.2 & \textbf{68.5} \\ 
& 1:2 & 56.5 & 97.1 & 71.4 & 61.1 \\ 
\hline 
 \multirow{2}{*}{\centering \textbf{tr}} 
& 1:1 & 89.3 & 50.1 & 64.2 & 71.6 \\ 
& 1:2 & 91.4 & 49.0 & 63.8 & \textbf{72.1} \\ 
\hline 
 \multirow{2}{*}{\centering \textbf{ar}} 
& 1:1 & 88.4 & 86.0 & 87.2 & 87.3 \\ 
& 1:2 & 86.1 & 91.0 & 88.5 & \textbf{88.1} \\ 
\hline 
 \multirow{2}{*}{\centering \textbf{ko}} 
& 1:1 & 80.6 & 67.5 & 73.4 & 72.0 \\ 
& 1:2 & 87.3 & 60.6 & 71.6 & \textbf{74.5} \\ 
\hline 
\multirow{2}{*}{\centering \textbf{Avg.}} 
&\cellcolor{gray!20}  1:1 &\cellcolor{gray!20} 74.4 &\cellcolor{gray!20} 79.0 &\cellcolor{gray!20} 73.4 &\cellcolor{gray!20} \textbf{69.7} \\ 
&\cellcolor{gray!20}  1:2 &\cellcolor{gray!20} 74.2 &\cellcolor{gray!20} 79.3 &\cellcolor{gray!20} 72.2 &\cellcolor{gray!20} 68.8 \\ 
\hline 
\hline
\end{tabular}
}

\end{table}

\begin{table}
\caption{We apply our proposed MHR to the CogVLM-chat model and evaluate on the POPE MUL benchmark. 
We use the ``popular" set. 
The average scores are marked in \colorbox{gray!20}{gray} and bold text denotes the best results.
}
\label{tab:cogvlm_pope}
\begin{tabular}{cc |ccc|c }

\hline
\textbf{Language}& {Method} &Precision & Recall & F1 Score& Acc. \\
\hline

\multicolumn{6}{c}{\textbf{High Resource Languages (HRL)}} \\   \hline 
 \multirow{2}{*}{\centering \textbf{en}} 
 & CogVLM & 91.1 & 79.1 & 84.7 & \textbf{85.7} \\ 
& Ours & 81.5 & 91.9 & 86.4 & 85.5 \\ 
\hline 
 \multirow{2}{*}{\centering \textbf{ru}} 
 & CogVLM & 45.6 & 100.0 & 62.6 & 20.0 \\ 
& Ours & 59.0 & 98.9 & 73.9 & \textbf{64.9} \\ 
\hline 
 \multirow{2}{*}{\centering \textbf{de}} 
 & CogVLM & 66.6 & 100.0 & 80.0 & 33.4 \\ 
& Ours & 99.6 & 88.5 & 93.8 & \textbf{36.8} \\ 
\hline 
 \multirow{2}{*}{\centering \textbf{zh}} 
 & CogVLM & 44.9 & 100.0 & 62.0 & 1.9 \\ 
& Ours & 54.6 & 100.0 & 70.7 & \textbf{41.2} \\ 
\hline 
 \multirow{2}{*}{\centering \textbf{ja}} 
 & CogVLM & 75.0 & 100.0 & 85.7 & 0.1 \\ 
& Ours & 88.5 & 100.0 & 93.9 & \textbf{27.1} \\ 
\hline 
 \multirow{2}{*}{\centering \textbf{fr}} 
 & CogVLM & 53.9 & 100.0 & 70.0 & 48.9 \\ 
& Ours & 84.5 & 99.4 & 91.4 & \textbf{66.1} \\ 
\hline 
 \multirow{2}{*}{\centering \textbf{es}} 
 & CogVLM & 92.3 & 72.2 & 81.0 & \textbf{81.8} \\ 
& Ours & 98.3 & 73.3 & 84.0 & 76.1 \\ 
\hline 
 \multirow{2}{*}{\centering \textbf{pt}} 
 & CogVLM & 50.0 & 100.0 & 66.7 & 50.0 \\ 
& Ours & 62.7 & 98.9 & 76.8 & \textbf{70.1} \\ 
\hline 
\multirow{2}{*}{\centering \textbf{Avg.}} 
&\cellcolor{gray!20}  CogVLM &\cellcolor{gray!20} 64.9 &\cellcolor{gray!20} 93.9 &\cellcolor{gray!20} 74.1 &\cellcolor{gray!20} 40.2 \\ 
&\cellcolor{gray!20}  Ours &\cellcolor{gray!20} 78.6 &\cellcolor{gray!20} 93.9 &\cellcolor{gray!20} 83.8 &\cellcolor{gray!20} \textbf{58.5} \\ 
\hline 
\multicolumn{6}{c}{\textbf{Low Resource Languages (LRL)}} \\  \hline 
 \multirow{2}{*}{\centering \textbf{uk}} 
 & CogVLM & 60.0 & 100.0 & 75.0 & 0.2 \\ 
& Ours & 62.9 & 95.6 & 75.9 & \textbf{59.1} \\ 
\hline 
 \multirow{2}{*}{\centering \textbf{bg}} 
 & CogVLM & 49.5 & 88.7 & 63.5 & 10.3 \\ 
& Ours & 62.3 & 96.1 & 75.6 & \textbf{68.9} \\ 
\hline 
 \multirow{2}{*}{\centering \textbf{tr}} 
 & CogVLM & 0.0 & 0.0 & 0.0 & 0.0 \\ 
& Ours & 80.4 & 26.6 & 40.0 & \textbf{56.8} \\ 
\hline 
 \multirow{2}{*}{\centering \textbf{ar}} 
 & CogVLM & 0.0 & 0.0 & 0.0 & 0.0 \\ 
& Ours & 81.9 & 44.0 & 57.2 & \textbf{67.1} \\ 
\hline 
 \multirow{2}{*}{\centering \textbf{ko}} 
 & CogVLM & 33.3 & 100.0 & 50.0 & 0.0 \\ 
& Ours & 88.8 & 100.0 & 94.1 & \textbf{11.7} \\ 
\hline 
\multirow{2}{*}{\centering \textbf{Avg.}} 
&\cellcolor{gray!20}  CogVLM &\cellcolor{gray!20} 28.6 &\cellcolor{gray!20} 57.7 &\cellcolor{gray!20} 37.7 &\cellcolor{gray!20} 2.1 \\ 
&\cellcolor{gray!20}  Ours &\cellcolor{gray!20} 75.3 &\cellcolor{gray!20} 72.5 &\cellcolor{gray!20} 68.5 &\cellcolor{gray!20} \textbf{52.7} \\ 
\hline 

\hline\hline

\hline
\end{tabular}

\end{table}

\subsection{Generality of MHR framework}
{In the main table, we have verified the effectiveness of our MHR framework on LLaVA 1.5, in this section, we further experiment with CogVLM to study the generality of our proposed MHR framework. 
CogVLM is a powerful open-source visual language foundation model that bridges the gap between a frozen pretrained language model and an image encoder through a trainable visual expert module integrated into the attention and FFN layers.
As shown in Table \ref{tab:cogvlm_pope}, with our proposed MHR, the CogVLM significantly improves the accuracy. Specifically, our framework improves the average accuracy of high-resource languages from 40.2 to 58.5. Notably, from this table, we can observe that the standard CogVLM has a very poor performance in low-resource languages. With our framework, the low-resource languages achieve more significant improvement. These experimental results comprehensively validate the generality of our framework.
}

\subsection{Qualitative Results}

To qualitatively verify the effectiveness of our method on downstream tasks, we presented two examples from the POPE MSCOCO dataset. As shown in Figure 7 (a), the samples are of the VQA type tasks and focus on the discriminative ability. Here we present six languages (en, uk, zh, fr, pt, and ja) to answer the input query. When querying LLaVA-1.5 with non-English languages, the model outputs wrong answers. In contrast, our model can accurately answer this question using all six languages. Moreover, to further demonstrate our model's capabilities in generation, we demonstrate two samples of image caption tasks. As denoted in Figure 7 (b), our model also presents significant better performance for mitigating hallucinations.

\section{Conclusion}

In this paper, we discover the severe multilingual hallucination problem in LVLMs and make the first attempt to alleviate it. Our empirical findings reveal that multilingual hallucinations can be attributed to two key factors. Based on them, we propose a two-stage Multilingual Hallucination Removal (MHR) framework, which first improves the instruction-following ability for different languages and then augments the model’s ability to resist hallucinations with a hallucination-enhanced
preference optimization. Experimental results demonstrate that our MHR achieves a substantial reduction in hallucination generation for LVLMs across both high-resource and low-resource languages. 

\begin{figure*}[!ht]
  \centering
  \subfloat[VQA Tasks]{
  \begin{minipage}[b]{0.48\textwidth}
    \includegraphics[width=1\textwidth]{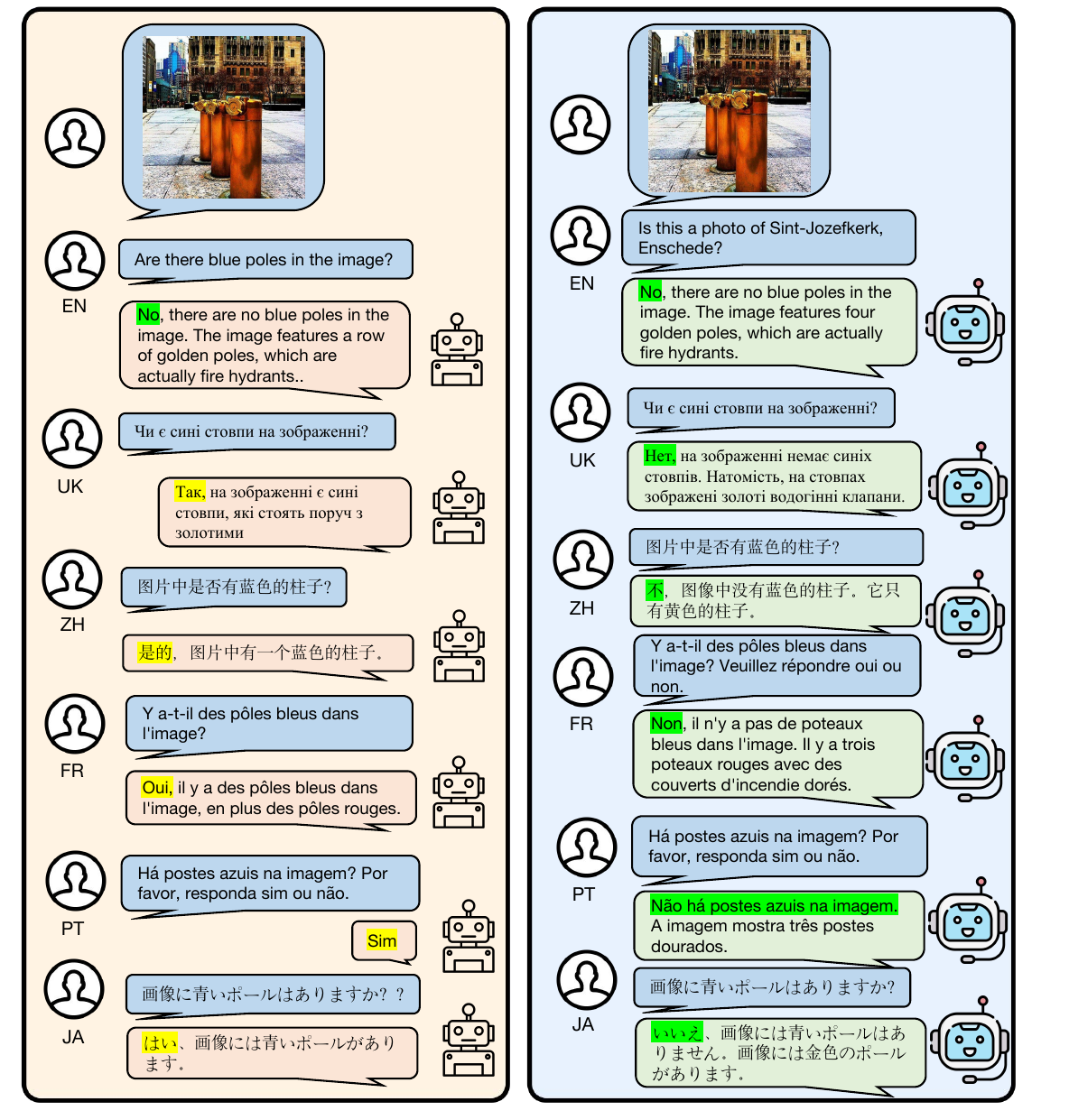} \\
    \includegraphics[width=1\textwidth]{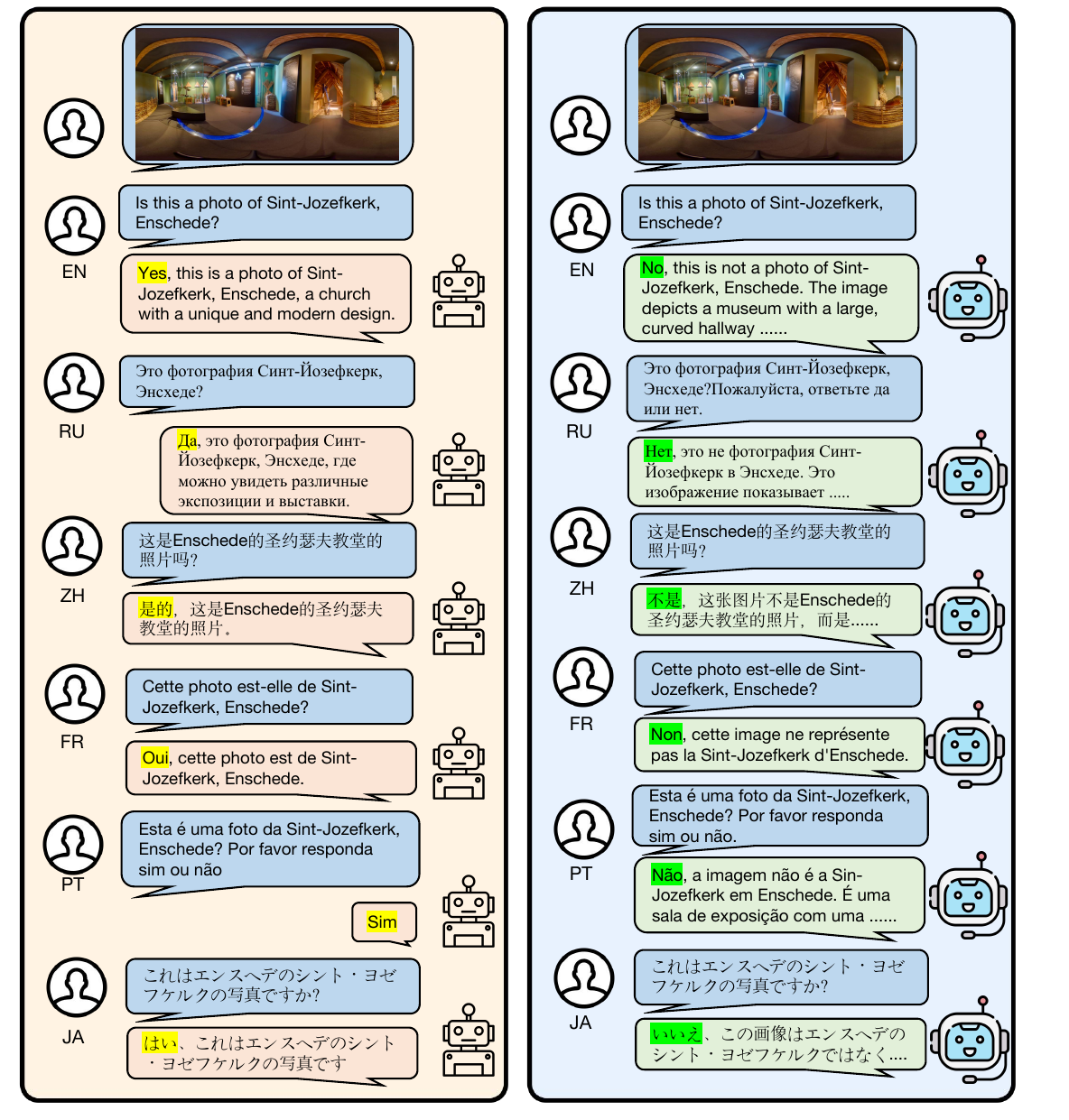} 
    \end{minipage}\
  }
  \hfil
  \subfloat[Image Captioning Tasks]{
  \begin{minipage}[b]{0.48\textwidth}
    \includegraphics[width=1\textwidth]{images/vqa1.pdf} \\
    \includegraphics[width=1\textwidth]{images/vqa2.pdf} 
    \end{minipage}\
  }
\caption{
Qualitative results of our proposed MHR method are presented. Figure (a) illustrates the results on VQA tasks, while Figure (b) provides a demonstration of the results pertaining to Image-Caption tasks. Responses generated by LLaVA 1.5 are exhibited within red containers, whereas answers originating from our final model are ensconced within green boxes. 
Hallucinated descriptions are highlighted in \colorbox{yellow!100}{yellow}, while descriptions void of hallucinations are marked in \colorbox{green!100}{green}. {Sentences rendered in red} represent those which are either not fluent or are not answered in the appropriate language.
}
\label{fig:quality12}
\end{figure*}

\bibliographystyle{IEEEtran}
\bibliography{main}

\end{document}